\documentclass{sigkddExp}

\usepackage{subfigure} 
\usepackage{amsfonts}
\usepackage{multirow}
\usepackage{threeparttable}
\usepackage{booktabs}

\begin{document}
%

\title{Adversarial Attacks and Defenses on Graphs:\\ A Review, A Tool and Empirical Studies}
%

\numberofauthors{1}
%


\author{
%

\alignauthor Wei Jin$^\dagger$, Yaxin Li$^\dagger$, Han Xu$^\dagger$, Yiqi Wang$^\dagger$, Shuiwang Ji$^\ddagger$, Charu Aggarwal$^\mathsection$, Jiliang Tang$^\dagger$\\
$^\dagger$\affaddr{Data Science and Engineering Lab, Michigan State University}\\
$^\ddagger$\affaddr{Texas A\&M University}\\
$^\mathsection$\affaddr{IBM T. J. Watson Research Center} \\
\email{\{jinwei2,liyaxin1,xuhan1,wangy206,tangjili\}@msu.edu, sji@tamu.edu, charu@us.ibm.com}


 }

\newcommand{\jw}[1]{\textcolor{blue}{@added or changed:~#1@}}
\newcommand{\jt}[1]{\textcolor{red}{@added or changed:~#1@}}

\maketitle

\begin{abstract}
Deep neural networks (DNNs) have achieved significant performance in various tasks. However, recent studies have shown that DNNs can be easily fooled by small perturbation on the input, called adversarial attacks. As the extensions of DNNs to graphs, Graph Neural Networks (GNNs) have been demonstrated to inherit this vulnerability. Adversary can mislead GNNs to give wrong predictions by modifying the graph structure such as manipulating a few edges. This vulnerability has arisen tremendous concerns for adapting GNNs in safety-critical applications and has attracted increasing research attention in recent years. Thus, it is necessary and timely to provide a comprehensive overview of existing graph adversarial attacks and the countermeasures. In this survey, we categorize existing attacks and defenses, and review the corresponding state-of-the-art methods. Furthermore, we have developed a repository with representative algorithms\footnote{{https://github.com/DSE-MSU/DeepRobust}}. The repository enables us to conduct empirical studies to deepen our understandings on attacks and defenses on graphs. 
\end{abstract}

\section{Introduction}

Graphs can be used as the representation of a large number of systems across various areas such as social science (social networks), natural science (physical systems, and protein-protein interaction networks), and knowledge graphs~\cite{aggarwal2010managing,hamilton2017representation}. With their prevalence, it is of great significance to learn effective representations for graphs that can facilitate various downstream tasks such as node classification, graph classification, link prediction and recommendation~\cite{aggarwal2010managing,hamilton2017representation}. Graph Neural Networks (GNNs), which generalize traditional deep neural networks (DNNs) to graphs, pave one effective way to learn representations for graphs~\cite{hamilton2017representation,wu2019comprehensive-survey,battaglia2018relational-survey,zhou2018graph-survey}. The power of GNNs relies on their capability of capturing the graph structure simultaneously with the node features. Instead of only considering the instances (nodes with their features) independently, GNNs also leverage the relationships between them. 
Specifically, GNNs follow a message-passing scheme~\cite{gilmer2017neural-mpnn}, where the nodes aggregate and transform the information from their neighbors in each layer. By stacking multiple GNN layers, information can be propagated further through the graph structure and we can embed nodes into low-dimensional representations. The obtained node representations can then be fed into any differentiable prediction layer so that the whole model can be trained in an end-to-end fashion. Due to their strong representation learning capability, GNNs
have gained practical significance in various applications ranging from data mining~\cite{kipf2016semi,gat}, natural language processing~\cite{marcheggiani2017encoding-nlp}, and computer vision~\cite{landrieu2018large-cv} to healthcare and biology~\cite{ma2018drug-healthcare}.

\begin{figure}
    \centering
    \includegraphics[width=0.9\linewidth]{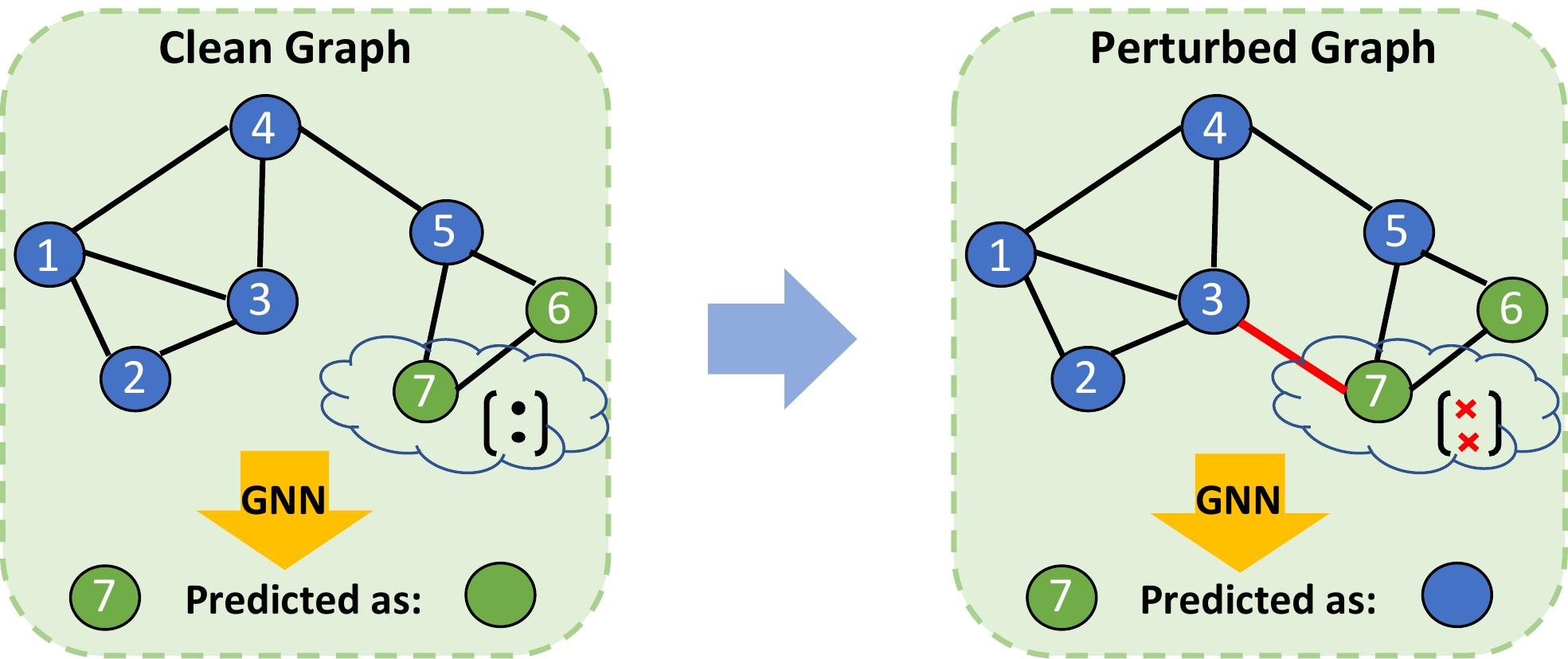}
    \caption{An example of adversarial attack on graph data. The goal of the GNN is to predict the color of the nodes. Here node 7 is the target node. Attacker aims to change the prediction of GNN on node 7 by modifying the edges and features.}
    \label{fig:attack_example}
\end{figure}

As new generalizations of traditional DNNs to graphs, GNNs inherit both advantages and disadvantages of traditional DNNs. Similar to traditional DNNs, GNNs are also powerful in learning representations of graphs and
have permeated numerous areas of science and technology. Traditional DNNs are easily fooled by adversarial attacks~\cite{goodfellow2014explaining,xu2019adversarial,li2020learning,li2019regional}. In other words, the adversary can insert slight perturbation during either the training or test phases, and the DNN models will totally fail. It is evident that GNNs also inherit this drawback~\cite{nettack,rl-s2v,metattack}. The attacker can generate graph adversarial perturbations by manipulating the graph structure or node features to fool the GNN models. As illustrated in Figure~\ref{fig:attack_example}, originally node $7$ was classified by the GNN model as a green node; after node $7$ creates a new connection with node $3$ and modifies its own features, the GNN model misclassifies it as a blue node. Such vulnerability of GNNs has arisen tremendous concerns on applying them in safety-critical applications such as financial system and risk management. For example, in a credit scoring system, fraudsters can fake connections with several high-credit customers to evade the fraud detection models; and  spammers can easily create fake followers to increase the chance of fake news being recommended and spread. Therefore, there is an urgent need to investigate graph adversarial attacks and their countermeasures.  

Pushing this research has a great potential to facilitate the successful adoption of GNNs in a broader range of fields, which encourages increasing attention on graph adversarial attacks and defenses in recent years. Thus, it is necessary and timely to provide a comprehensive and systematic overview on existing algorithms. Meanwhile, it is of great importance to deepen our understandings on graph adversarial attacks via empirical studies. These understandings can not only provide knowledge about the behaviors of attacks but also offer insights for us to design defense strategies. These motivate this survey with the following key purposes: 
\begin{itemize}
    \item We categorize existing attack methods from various perspectives in Section 3 and review representative algorithms in Section 4.
    \item We classify existing countermeasures according to their defense strategies and give a review on representative algorithms for each category in Section 5.  
    \item We perform empirical studies based on the repository we developed that provide comprehensive understandings on graph attacks and defenses in Section 6.     
    \item We discuss some promising future research directions in Section 7. 
\end{itemize}


\section{Preliminaries and Definitions}
Before presenting the review and empirical studies, we first introduce concepts, notations and definitions in this section. 
 
\subsection{Learning on Graph Data}\label{gnn}
In this survey, we use $G=(V,E)$ to denote the structure of a graph where $V=\{v_1, \dots{}, v_N\}$ is the set of $N$ nodes and $E=\{e_1, \dots{}, e_K\}$ is the edge set.  We use matrix ${\bf A}\in \{0, 1\}^{{N}\times{N}}$ to denote the adjacency matrix of $G$, where each entry ${\bf A}_{ij} = 1$ means nodes $v_i$ and $v_j$ are connected in $G$. Furthermore, we use ${\bf X}\in\mathbb{R}^{{N}\times{D}}$ to denote the node attribute matrix where $D$ is the dimension of the node feature vectors. Thus, graph data can be denoted as $G=({\bf A}, {\bf X})$. There are a lot of learning tasks on graphs and in this work, we focus on the classification problems on graphs. Furthermore, we use $f_\theta$ with parameters $\theta$ to denote the learning models in this survey.



\vskip 0.5em

\noindent{}\textbf{Node-Level Classification.} 
For node-level classification, each node in the graph $G$ belongs to a class in the label set $Y$. The graph model aims to learn a neural network, based on labeled nodes (training nodes), denoted as $V_L$, to predict the class of unlabeled nodes (test nodes). The training objective function can be formulated as:  
\begin{equation}
\min _{\theta}~ \mathcal{L}_{\text{train}}(f_\theta(G))=\sum_{v_i \in {V}_{L}} \ell\left(f_{\theta}({\bf X}, {\bf A})_{i}, y_{i}\right),
\end{equation}
where $f_\theta( {\bf X}, {\bf A})_i$ and $y_i$ are the predicted and the true label of node $v_i$ and $\ell(\cdot, \cdot)$ is a loss function such as cross entropy.

%

\vskip 0.5em
\noindent{}\textbf{Graph-Level Classification.} 
For graph-level classification, each individual graph has a class in the label set $Y$. We use $\mathcal{G}$ to denote a set of graphs, and $\mathcal{G}_L$ is the labeled set (training set) of $\mathcal{G}$. The goal of graph-level classification is to learn a mapping function $f_\theta: \mathcal{G}\rightarrow{Y}$ to predict the labels of unlabeled graphs. Similar to node-level classification, the objective function can be formulated as  
\begin{equation}
\min _{\theta}~ \mathcal{L}_{\text{train}}(\mathcal{G})=\sum_{G_i \in \mathcal{G}_{L}} \ell\left(f_{\theta}(G_i), y_{i}\right),
\end{equation}
where $G_i$ is the labeled graph with ground truth $y_i$ and $f_\theta(G_i)$ is the prediction of the graph $G_i$. 

\subsection{A General Form of Graph Adversarial Attack}
Based on the objectives in Section~\ref{gnn}, we can define a general form of the objective for adversarial attacks, which aims to  misguide the victim model by minimizing some attack loss. Thus, the problem of node-level graph adversarial attacks can be stated as:
\vskip 0.5em

\textit{Given $G=({\bf A}, {\bf X})$ and victim nodes subset $V_t\subseteq{V}$. Let $y_u$ denote the class for node $u$ (predicted or using ground truth). The goal of the attacker is to find a perturbed graph $\hat{G}=({\bf \hat{A}, \hat{X}})$ that minimize the following attack objective} $\mathcal{L}_{\text{atk}}$,
\begin{equation}
\label{eq:unified}
\begin{aligned}
& {\min~ \mathcal{L}_{\text{atk}}(f_\theta({\hat{G}})) = \sum_{u \in V_{t}} \ell_{\text{atk}}\left(f_{\theta^{*}}(\hat{G})_u, y_{u}\right)} \\
& s.t., \quad {\theta^{*}=\arg \min _{\theta} \mathcal{L}_{\text{train}}\left(f_{\theta}\left({G^\prime}\right)\right)}, \\
\end{aligned}
\end{equation}
\textit{where $\ell_{\text{atk}}$ is a loss function for attacking and one option is to choose $\ell_{\text{atk}}=-\ell$ and $G^\prime$ can either be $G$ or $\hat{G}$. Note that $\hat{G}$ is chosen from a constrained domain $\Phi(G)$. Given a fixed perturbation budget $\Delta$, a typical $\Phi(G)$ can be implemented as,}
\begin{equation}
\|\hat{\bf A}-{\bf A}\|_{0}+\|\hat{\bf X}-{\bf X}\|_{0}\leq\Delta.
\end{equation}

\begin{table*}[t]
    \small
    \centering
    \caption{Commonly used notations}
    \label{tab:notations}
    \begin{tabular}{c|c|c|c}
    \toprule
        Notations & Description & Notations & Description \\
    \midrule
         $G$ &  Graph & $u$ & Target node\\
         $\hat{G}$ &  Perturbed graph & $y_u$ & Label of node $u$\\
         $V$ &  The set of nodes & $f_{\theta}$ & Neural network model\\
         $V_L$ &  The set of labeled nodes & $\mathcal{L}$ & Loss function \\
         $E$ &  The set of edges & $l(\cdot,\cdot)$ & Pair-wise loss function\\
         $\bold{A}$ & Adjacency matrix & $\|\cdot\|_{0}$ & $\ell_0$ norm  \\
         $\bold{\hat{A}}$ & Perturbed adjacency matrix & $\Delta $ & Perturbation budget\\
         $\bold{X}$ & Node attribute matrix & ${\bf Z}$ & Predicted probability\\
         $\bold{\hat{X}}$ & Perturbed node attribute matrix & $\bold{h}_u $ & Hidden representation of node $u$\\
         $D $ & Dimension of node features & 
         $e_{ij}$ & Edge between node $v_i$ and $v_j$\\

    \bottomrule
    \end{tabular}
    
    \label{tab:notations}
\end{table*}

We omit the definition of graph-level adversarial attacks since (1) the graph-level adversarial attacks can be defined similarly and (2) the majority of the adversarial attacks and defenses focus on node-level. Though adversarial attacks have been extensively studied in the image domain, we still need dedicated efforts for graphs due to unique challenges -- (1) graph structure is discrete; (2) the nodes in the graph are not independent; and (3) it is difficult to measure whether the perturbation on the graph is imperceptible or not.

\subsection{Notations}
With the aforementioned definitions, we list all the notations which will be used in the following sections in Table~\ref{tab:notations}.


\section{Taxonomy of Graph Adversarial Attacks}

In this section, we briefly introduce the main taxonomy of adversarial attacks on graph structured data. Attack algorithms can be categorized into different types based on different goals, resources, knowledge and capacity of attackers. We try to give a clear overview on the main components of graph adversarial attacks.  

\subsection{Attacker's Capacity}
The adversarial attacks can happen at two phases, i.e., the model training and  model testing. It depends on the attacker's capacity to insert adversarial perturbation:
\begin{itemize}
    \item \textbf{Evasion Attack:} Attacking happens after the GNN model is trained or in the test phase. The model is fixed, and the attacker cannot change the model parameter or structure. The attacker performs evasion attack when $G^\prime=G$ in Eq.~(\ref{eq:unified}).
    \item \textbf{Poisoning Attack:} Attacking happens before the GNN model is trained. The attacker can add ``poisons'' into the model training data, letting trained model have malfunctions. It is the case when $G^\prime=\hat{G}$ in Eq.~(\ref{eq:unified}).
\end{itemize}

\subsection{Perturbation Type}
The attacker can insert adversarial perturbations from different aspects. The perturbations can be categorized as modifying node features, adding/deleting edges, and adding fake nodes. Attackers should also keep the perturbation unnoticeable, otherwise it would be easily detected.
\begin{itemize}
    \item {\bf Modifying Feature:} Attackers can slightly change the node features while maintaining the graph structure.
    \item {\bf Adding or Deleting Edges:} Attackers can add or delete edges for the existing nodes under certain budget of total actions. 
    \item {\bf Injecting Nodes:} Attackers can insert fake nodes to the graph, and link it with some benign nodes in the graph.
\end{itemize}

\subsection{Attacker's Goal}

According to the goals of attacks, we can divide the attacks for node-level classification into the following two categories:
\begin{itemize}
    \item \textbf{Targeted Attack:} There is a small set of test nodes. The attacker aims to let the trained model misclassify these test samples. It is the case when $V_t\subset{V}$ in Eq.~(\ref{eq:unified}). We can further divide targeted attacks into (1) direct attack where the attacker directly modifies the features or edges of the target nodes and (2) influencer attack where the attacker can only manipulate other nodes to influence the targets.  
    \item \textbf{Untargeted Attack:} The attacker aims to insert poisons to let the trained model have bad overall  performance on all test data. It is the case when $V_t={V}$ in Eq.~(\ref{eq:unified}).  
\end{itemize}
Noth that for graph-level classification, there also exist targeted and untargeted attacks. Targeted attack aims to induce the model to give a specific label to a given graph sample, while untargeted attack only wants the model to perform incorrectly.

\subsection{Attacker's Knowledge}
Attacker's knowledge means how much information an attacker knows about the model that he aims to attack. Usually, there are three settings:
\begin{itemize}
    \item \textbf{White-box Attack:} All information about the model parameters, training input (e.g, adjacency matrix and attribute matrix) and the labels are given to the attacker.   
    \item \textbf{Gray-box Attack:} The attacker only has limited knowledge about the victim model. For example, the attacker cannot access the model parameters but can access the training labels. Then it can utilize the training data to train surrogate models to estimate the information from victim model. 
    \item \textbf{Black-box Attack:} The attacker does not have access to the model's parameters or training labels. It can access the adjacency matrix and attribute matrix, and do black-box query for output scores or labels. 
\end{itemize}

\subsection{Victim Models}

In this part we are going to summarize the victim models that have been proven to be susceptible to adversarial examples. 

\vskip 0.5em
\noindent{}\textbf{Graph Neural Networks.} 
Graph neural networks are powerful tools in learning representation of graphs~\cite{graph-attack-survey-2018,wu2019comprehensive-survey}. 
One of the most successful GNN variants is Graph Convolutional Networks (GCN)~\cite{kipf2016semi}.  GCN learns the representation for each node by keeping aggregating and transforming the information from its neighbor nodes. Though GNNs can achieve high performance in various tasks, studies have demonstrated that GNNs including GCN are vulnerable to adversarial attacks~\cite{nettack,graph-attack-survey-2018}. Besides, it is evident from recent works~\cite{nettack,rgcn,ma2020black} that other graph neural networks including column network (CLN)~\cite{pham2016column}, graph attention network (GAT)~\cite{gat} and JKNet~\cite{xu2018representation-jknet} have the same issue.

\vskip 0.5em
\noindent{}\textbf{Other Graph Learning Algorithms.} In addition to graph neural networks, adversary may attack some other important algorithms for graphs such as network embeddings including LINE~\cite{tang2015line} and Deepwalk~\cite{perozzi2014deepwalk}, graph-based semi-supervised learning (G-SSL)~\cite{label-learning-propagation}, and knowledge graph embedding ~\cite{bordes2013translating-transE,lin2015learning-transR}.

\begin{table*}[t]
\centering
\caption{Categorization of representative attack methods}
\label{tab:Categorization}
\resizebox{0.8\textwidth}{!}{
\begin{tabular}{@{}c|c|c|c|c|c|c@{}}
\toprule
Attack Methods                                               & \begin{tabular}[c]{@{}l@{}}Attack \\ Knowledge\end{tabular}  & \begin{tabular}[c]{@{}l@{}}Targeted or\\ Non-targeted\end{tabular} & \begin{tabular}[c]{@{}l@{}}Evasion or\\ Poisoning\end{tabular} & Perturbation Type                                                          & Application                                                                        & Victim Model                                                            \\ \midrule
PGD, Min-max~\cite{xu2019topology-attack}                & White-box                                                    & Untargeted                                                         & Both                                                           & Add/Delete edges                                                           & Node Classification                                                                & GNN                                                                     \\
\midrule \begin{tabular}[c]{@{}c@{}}IG-FGSM~\cite{deep-insight-jaccard} \\ IG-JSMA~\cite{deep-insight-jaccard}\end{tabular} & White-box                                                    & Both                                                               & Evasion                                                        & \begin{tabular}[c]{@{}c@{}}Add/Delete edges\\ Modify features\end{tabular}                            & Node Classification                                                                & GNN                                                                     \\

\midrule Wang et al.~\cite{wang2019attacking-ccs}                & \begin{tabular}[c]{@{}l@{}}White-box\\ Gray-box\end{tabular} & Targeted                                                           & Poisoning                                                      & Add/Delete edges                                                           & Node Classification                                                                & GNN                                                                     \\
\midrule Nettack~\cite{nettack}                              & Gray-box                                                     & Targeted                                                           & Both                                                           & \begin{tabular}[c]{@{}c@{}}Add/Delete edges\\ Modify features\end{tabular} & Node Classification                                                                & GNN                                                                     \\
\midrule Metattack~\cite{metattack}                            & Gray-box                                                     & Untargeted                                                         & Poisoning                                                      & Add/Delete edges                                                           & Node Classification                                                                & GNN                                                                     \\
\midrule NIPA~\cite{nipa}                                 & Gray-box                                                    & Untargeted                                                         & Poisoning                                                      & Inject nodes                                                               & Node Classification                                                                & GNN                                                                     \\
\midrule RL-S2V~\cite{rl-s2v}                               & Black-box                                                    & Targeted                                                           & Evasion                                                        & Add/Delete edges                                                           & \begin{tabular}[c]{@{}l@{}}Graph Classification\\ Node Classification\end{tabular} & GNN                                                                     \\
\midrule ReWatt~\cite{rewatt}                               & Black-box                                                    & Untargeted                                                         & Evasion                                                        & Add/Delete edges                                                           & Graph Classification                                                               & GNN                                                                     \\

\midrule Liu et al.~\cite{liu2019unified-ssl}                   & \begin{tabular}[c]{@{}l@{}}White-box\\ Gray-box\end{tabular} & Untargted                                                          & Poisoning                                                      &\begin{tabular}[c]{@{}c@{}}Flip label\\ Modify features\end{tabular}                                                                  & \begin{tabular}[c]{@{}l@{}}Classification\\ Regression\end{tabular}                & G-SSL                                                                   \\

\midrule FGA~\cite{chen2018fast-gradient-network-embedding}                            & White-box                                                    & Targeted                                                           & Both                                                      & Add/Delete edges                                                           & \begin{tabular}[c]{@{}c@{}}Node Classification\\ Community Detection\end{tabular}                                                                 & \begin{tabular}[c]{@{}c@{}}Network \\ Emebdding\end{tabular}            \\
\midrule GF-Attack~\cite{gf-attack}                            & Black-box                                                    & Targeted                                                           & Evasion                                                      & Add/Delete edges                                                           & Node Classification                                                                & \begin{tabular}[c]{@{}c@{}}Network \\ Emebdding\end{tabular}            \\
\midrule Bojchevski et al.~\cite{node-embedding-poisoning}             & Black-box                                                    & Both                                                           & Poisoning                                                      & Add/Delete edges                                                           & \begin{tabular}[c]{@{}c@{}}Node Classification\\ Community Detection\end{tabular}  & \begin{tabular}[c]{@{}c@{}}Network \\ Emebdding\end{tabular}            \\

\midrule Zhang et al.~\cite{TowardsDP-knowledge-graph}             & White-box                                                    & Targeted                                                           & Poisoning                                                      & Add/Delete facts                                                           & Plausibility Prediction                                                            & \begin{tabular}[c]{@{}c@{}}Knowledge \\ Graph \\ Embedding\end{tabular} \\
\midrule CD-Attack~\cite{li2020adversarial-hiding-community} & Black-box                                                    & Targeted                                                           & Poisoning                                                      & Add/Delete edges                                                           & Community Detection                                                                & \begin{tabular}[c]{@{}c@{}}Community \\ Detection \\ Algorithm\end{tabular} \\ 
\bottomrule
\end{tabular}}
\end{table*}


\section{ Graph Adversarial Attacks}
In this section, we review representative algorithms for graph adversarial attacks. Following the categorizations in the previous section, we first divide these algorithms into white-box, gray-box and black-box and then for algorithms in each category, we further group them into targeted and untargeted attacks. Note that without specific mention, the following attack methods focus on node-level classification. An overall categorization of representative attack methods is shown in Table~\ref{tab:Categorization}. In addition, some open source implementations of representative algorithms are listed in Table~\ref{tab:code}.

\subsection{White-box Attacks}
In white-box attack setting, the adversary has access to any information about the victim model such as model parameters, training data, labels, and predictions. Although in most of the real world cases we do not have the access to such information, we can still assess the vulnerability of the victim models under the worst situation. Typically, white-box attacks use the gradient information from the victim model to guide the generation of attacks~\cite{chen2018fast-gradient-network-embedding,xu2019topology-attack,deep-insight-jaccard,chen2019ga-attack-community}. 

\subsubsection{Targeted Attack}

Targeted attack aims to mislead the victim model to make wrong predictions on some target samples. A lot of studies follow the white-box targeted attack setting with a wide range of real-world applications. FGA~\cite{chen2018fast-gradient-network-embedding} extracts the link gradient information from GCN, and then greedily selects the pair of nodes with maximum absolute gradient to modify the graph iteratively. Genetic algorithm based Q-Attack is proposed to attack a number of community detection algorithms~\cite{chen2019ga-attack-community}. Iterative gradient attack (IGA) based
on the gradient information in the trained graph auto-encoder, which is introduced to attack link prediction~\cite{chen2018link-prediction}. Furthermore, the vulnerability of knowledge graph embedding is investigated in~\cite{TowardsDP-knowledge-graph} and the plausibility of arbitrary facts in knowledge graph can be effectively manipulated by the attacker. Recommender systems based on GNNs are also vulnerable to adversarial attacks, which is shown in~\cite{attack-gcmc-recommender}. In addition, there are great efforts on attacking node-level classification. 
Traditional attacks in the image domain always use models' gradients to find adversarial examples. However, due to the discrete property of graph data, directly calculating gradients of models could fail. To solve this issue, the work~\cite{deep-insight-jaccard} suggests to use integrated gradient~\cite{sundararajan2017axiomatic} to better search for adversarial edges and feature perturbations. During the attacking process, the attacker iteratively chooses the edge or feature which has the strongest effect to the adversarial objective. By this way, it can cause the victim model to misclassify target nodes with a higher successful rate. The work~\cite{zang2020graph-universal-attack} assumes there is a set of ``bad actor'' nodes in a graph. When they flip the edges with any target nodes in a graph, it will cause the GNN model to have a wrong prediction on the target node. These ``bad actor'' nodes are critical to the safety of GNN models. For example, Wikipedia has hoax articles which have few and random connections to real articles. Manipulating the connections of these hoax articles will cause the system to make wrong prediction of the categories of real articles.



\subsubsection{Untargeted Attack}
Currently there are not many studies on untargeted white-box attack, and topology attack~\cite{xu2019topology-attack} is one representative algorithm. It first constructs a binary symmetric perturbation matrix ${\bf S}\in\{0,1\}^n$ where ${\bf S}_{ij}=1$ indicates to flip the edge between $i$ and $j$ and ${\bf S}_{ij}=0$ means no modification on ${\bf A}_{ij}$. Thus, the goal of the attacker is to find ${\bf S}$ that minimizes the predefined attack loss given a finite budget of edge perturbations $\Delta$, i.e., $\|{\bf S}\|_{0}\leq \Delta$. It considers two different attack scenarios: attacking pre-trained GNN with fixed parameters $\theta$ and attacking a re-trainable GNN $f_\theta$. For attacking a fixed $f_\theta$, the problem can be formulated as, 
\begin{equation}
    \begin{aligned}
        & \min\limits_{{\bf{S}}\in{\{0,1\}^n}}~\mathcal{L}_{\text{atk}}(f_\theta(\bf{S, {A}},\bf{X}))
        & \quad \text{s.t.}~ \|{\bf S}\|_{0}\leq{\Delta}.
    \end{aligned}
\end{equation}
It utilizes the Projected Gradient Descent (PGD) algorithm in~\cite{madry2017towards} to search the optimal $\bf S$. Note that the work \cite{madry2017towards} is also one popular attack algorithm in the image domain. For the re-trainable GNNs, parameter $\theta$ will be retrained after adversarial manipulation. Thus the attack problem is formulated as a min-max form where the inner maximization is to update $\theta$ by maximizing the attack loss and can be solved by gradient ascent while the outer minimization can be solved by PGD. Another work on untargeted white-box attack aims to attack multi-network
mining by perturbing the networks (graphs)~\cite{zhou2019admiring}. Specifically, in each iteration it measures the influence of network elements (edges and node attributes) to the mining results and attack one network element with the highest influence value.

\subsection{Gray-box Attacks}

White-box attacks assume that attackers can calculate gradient through model parameters, which is not always practical in real-world scenarios. Gray-box attacks are proposed to generate attacks with limited knowledge on the victim model~\cite{nettack,metattack,nipa}. Usually they first train a surrogate model with the labeled training data to approximate the information of the victim model and then generate perturbations to attack the surrogate model. It is noted that these models need the access to the labels of training data, thus they are not black-box attacks that will be introduced in the following subsection.

\subsubsection{Targeted Attack}
The early work on targeted gray-box attacks is for graph clustering~\cite{chen2017practical-clustering}. It demonstrates that injecting noise to a domain name system (DNS) query graph can degrade the performance of graph embedding models. Different from ~\cite{chen2017practical-clustering}, the work~\cite{nettack} proposes an attack method called Nettack to generate structure and feature attacks, aiming at solving Eq.~(\ref{eq:unified}). Besides, they argue that only limiting the perturbation budgets cannot always make the perturbation ``unnoticeable''. They suggest the perturbed graphs should also maintain important graph properties, including degree distribution and feature co-occurrence. Therefore, Nettack first selects possible perturbation candidates not violating degree distribution and feature co-occurrence of the original graph. Then it greedily chooses the perturbation that has the largest score to modify the graph, where the score is defined as,
\begin{equation}
\max _{i \neq y} \ln \left({\bf Z}_{u,i}\left( G^{\prime}\right)\right)-\ln \left({\bf Z}_{u,y}\left( G^{\prime}\right)\right),
\end{equation}
where ${\bf Z}_{u,i}$ is the probability of node $u$ to be the class $i$ predicted by the surrogate model. Thus, the goal of the attacker is to maximize the difference in the log-probabilities
of the target node $u$. By doing this repeatedly until reaching the perturbation budge $\Delta$, it can get the final modified graph. Furthermore, it suggests that such graph attack can also transfer from model to model, just as the attacks in the image domain~\cite{goodfellow2014explaining}. The authors also conduct influencer attacks where they can only manipulate the nodes except the target. It turns out that influencer attacks lead to a lower decrease in performance compared with directly modifying target node given the same perturbation budget.

However, Nettack cannot handle large-scale graphs due to its high time complexity. The work~\cite{zugner2020adversarial-tkdd} employs statistical analysis to investigate the patterns exhibited by Nettack perturbations. According to the observations on perturbation patterns, a scalable attack method Fasttack is proposed. It first ranks the possible perturbations by their impact on the victim model and chooses the top perturbations to generate attacks. In~\cite{wang2020scalable}, AFGSM is proposed to derive an approximate closed-form solution with a lower time cost for the attack model and sequentially inject fake nodes into the graph. 

Another line of works focus on generating backdoor attacks for graphs~\cite{zhang2020backdoor,xi2020graph-backdoor}. Backdoor attack aims to find some "trigger" (or hidden pattern) that would produce
misclassified results when added to an input~\cite{wang2019neural-backdoor, chen2017targeted}. To generate backdoor attacks for graphs, the attacker injects triggers (e.g. carefully-crafted subgraphs) in the training samples, and aims to fool GNNs into misclassifying those samples with triggers while maintaining the performance on other testing samples. Note that graph backdoor attacks can be applied in both node-level and graph-level classification~\cite{xi2020graph-backdoor}.

\subsubsection{Untargeted Attack}
Although following the same way of training a surrogate model as Nettack, Metattack~\cite{metattack} is a kind of untargeted poisoning attack. It tackles the bi-level problem in Eq.~(\ref{eq:unified}) by using meta-gradient. Basically, it treats the graph structure matrix as a hyper-parameter and the gradient of the attacker loss with respect to it can be obtained by:
\begin{align}
& \nabla_{G}^{\text {meta }}=\nabla_{G} \mathcal{L}_{\text {atk }}\left(f_{\theta^{*}}(G)\right).
\end{align}
Note that $\nabla_{G} \mathcal{L}_{\text {atk }}\left(f_{\theta^{*}}(G)\right)$ is actually a function with respect to both $G$ and $\theta$. If $\theta^*$ is obtained by some differential operations, we can compute $\nabla_{G}^{\text {meta }}$ as follows,
\begin{equation}
\nabla_{G}^{\text {meta }}= \nabla_{f} \mathcal{L}_{\text {atk }} \left(f_{\theta^{*}}(G)\right) \cdot\left[\nabla_{G} f_{\theta^{*}}(G)+\nabla_{\theta^{*}} f_{\theta^{*}}(G) \cdot \nabla_{G} \theta^{*}\right]
\end{equation}
where $\theta^*$ is often obtained by gradient descent in fixed iterations $T$. At iteration $t+1$, the gradient of $\theta_{t+1}$ with respect to $G$ can be formulated as,  
\begin{equation}
\nabla_{G} \theta_{t+1}=\nabla_{G} \theta_{t}-\alpha \nabla_{G} \nabla_{\theta_{t}} \mathcal{L}_{\text {train }}\left(f_{\theta_{t}}(G)\right),
\end{equation}
where $\alpha$ denotes learning rate of the gradient descent operation. By unrolling the training procedure from $\theta_T$ back to $\theta_0$, we can get $\nabla_{G}\theta_T$ and then $\nabla_{G}^{\text {meta }}$.  A greedy approach is applied to select the perturbation based on the meta gradient. Specifically, given the meta-gradient for a node pair $(u,v)$, it defines a score $S(u,v)=\nabla_{{\bf A}_{u v}}^{\operatorname{meta}}\cdot\left(-2 \cdot {\bf A}_{uv}+1\right)$ and greedily picks the perturbation which has the highest score but satisfies the unnoticeable constraints as in Nettack.

Instead of modifying the connectivity of existing nodes, a novel reinforment learning method for node injection poisoning attacks (NIPA)~\cite{nipa} is proposed to inject fake nodes into graph data . Specifically, NIPA first injects singleton $n$ nodes into the original graph. Then in each action $a_t$, the attacker first chooses an injected node to connect with another node in the graph and then assigns a label to the injected node. By doing this sequentially, the final graph is statistically similar to the original graph but can degrade the overall model performance.

\subsection{Black-box Attacks}

Different from gray-box attacks, black-box attacks~\cite{rl-s2v,rewatt,nipa,node-embedding-poisoning,gf-attack, ma2020black} are more challenging since the attacker can only access the adjacency matrix, attribute matrix and output of the victim model. The access of parameters, labels and predicted probability is prohibited. 
\subsubsection{Targeted Attack}
As mentioned earlier, training a surrogate model requires access to the labels of training data, which is not always practical. We hope to find a way that we only need to do black-box query on the victim model~\cite{rl-s2v} or attack the victim in an unsupervised fashion~\cite{node-embedding-poisoning,gf-attack}. 

To do black-box query on the victim model, reinforcement learning is introduced. RL-S2V~\cite{rl-s2v} is the first work to employ reinforcement learning technique to generate adversarial attacks on graph data under the black-box setting. They model the attack procedure as a Markov Decision Process (MDP) and the attacker is allowed to modify $m$ edges to change the predicted label of the target node $u$. They study both node-level (targeted) and graph-level (untargeted) attacks. For node-level attack, they define the MDP as follows, 
\begin{itemize}
    \item \textbf{State.} The state $s_{t}$ is represented by the tuple $\left(G^{(t)}, u\right)$ where $G^{(t)}$ is the modified graph at time step $t$.
    \item \textbf{Action.} A single action at time step $t$ is denoted as $a_t$. For each action $a_t$, the attacker can choose to add or remove an edge from the graph. Furthermore,  a hierarchical structure is applied to decompose the action space. 
    \item \textbf{Reward.} Since the goal of the attacker is to change the classification result of the target node $u$, RL-S2V gives non-zero reward $r$ to the attacker at the end of the MDP: 
    \begin{equation}
        r\left(s_{m}, a_{m}\right)=\left\{\begin{array}{ll}
    {1} & {\text { if } f_\theta\left(G^{(m)}, u\right) \neq y}, \\
    {-1} & {\text { if } f_\theta\left(G^{(m)}, u\right)=y.}
    \end{array}\right. 
    \end{equation}
    In the intermediate steps, the attacker receives no reward, i.e., $\forall t=1,2, \dots, m-1, r\left(s_{t}, a_{t}\right)=0$. 
\item \textbf{Termination.} The process terminates when the attacker finishes modifying $m$ edges.
\end{itemize}
Since they define the MDP of graph-level attack in the similar way, we omit the details. Further, the Q-learning algorithm~\cite{mnih2013playing} is adopted to solve the MDP and guide the attacker to modify the graph. 

Instead of attacking node-level classification, the work~\cite{node-embedding-poisoning} shows a way to attack the family of node embedding models in the black-box setting. Inspired by the observation that DeepWalk can be formulated in matrix factorization form~\cite{qiu2018network}, they maximize the unsupervised DeepWalk loss with matrix perturbation theory by performing $\Delta$ edge flips. It is further demonstrated that the perturbed structure is transferable to other models like GCN and Label Propagation.
However, this method only considers the structure information. {GF-Attack}~\cite{gf-attack} is proposed to incorporate the feature information into the attack model. Specifically, they formulate the connection between the graph embedding method and general graph signal process with graph filter and construct the attacker based on the graph filter and attribute matrix. {GF-Attack} can also be transferred to other network embedding models and achieves better performance than the method in ~\cite{node-embedding-poisoning}.  

\subsubsection{Untargeted Attack}
It is argued that the perturbation constraining only the number of modified edges may not be unnoticeable enough. A novel framework ReWatt~\cite{rewatt} is proposed to solve this problem and perform untargeted graph-level attack. Employing a reinforcement learning framework, ReWatt adopts the rewiring operation instead of simply adding/deleting an edge in one single modification to make perturbation more unnoticeable. One rewiring operation involves three nodes $v_1, v_2$ and $v_3$, where ReWatt removes the existing edge between $v_1$ and $v_2$ and connects $v_1$ and $v_3$. ReWatt also constrains $v_3$ to be the 2-hop neighbor of $v_1$ to make perturbation smaller. Such rewiring operation does not change the number of nodes and edges in the graph and it is further proved that such rewiring operation affects algebraic connectivity and effective graph resistance, both of which are important graph properties based on graph Laplacian, in a smaller way than adding/deleting edges. 

To perform graph adversarial attacks in a more realistic way, the work~\cite{ma2020black} proposes a more restricted black-box setting where the attackers only have access to a subset of nodes and can only attack a small number of them. Under this setting, the attacker is required to generate attacks in two steps: 1) selecting a small subset of nodes to attack under the limits of node access; 2) modifying node features or edges under the perturbation budget. By exploiting the structural inductive biases of the GNN models~\cite{xu2018representation-jknet,klicpera2018predict-appnp} as the information source for attacking, it proposes a practical greedy method of adversarial attacks for node-level classification tasks and effectively degrades the performance of GNNs.

\section{Countermeasures Against \\ Graph Adversarial Attacks}
In previous sections, we have shown that graph neural networks can be easily fooled by unnoticeable perturbation on graph data. The vulnerability of graph neural networks poses great challenges to apply them in safety-critical applications. In order to defend the graph neural networks against these attacks, different countermeasure strategies have been proposed. The existing methods can be categorized into the following types: (1) adversarial training, (2) adversarial perturbation detection, (3) certifiable robustness, (4) graph purification,  and (5) attention mechanism. 

\subsection{Adversarial Training}
Adversarial training is a widely used countermeasure for adversarial attacks in image data~\cite{goodfellow2014explaining}. The main idea of adversarial training is to inject adversarial examples into the training set such that the trained model can correctly classify the future adversarial examples.  Similarly, we can also adopt this strategy to defend graph adversarial attacks as follows, 
\begin{equation}
\min _{\theta} \max _{\delta_{\bf A} \in \mathcal{P}_{\bf A} \atop \delta_{\bf X} \in \mathcal{P}_{\bf X}} \mathcal{L}_{\text{train}}\left(f_{\theta}({\bf A}+\delta_{\bf A}, {\bf X}+\delta_{\bf X}) \right),
\label{eq:adv-training}
\end{equation}
where $\delta_{\bf A}$, $\delta_{\bf X}$ denote the perturbation on ${\bf A}, {\bf X}$, respectively; $\mathcal{P}_{\bf A}$ and $\mathcal{P}_{\bf X}$ stand for the domains of imperceptible perturbation. The min-max optimization problem in Eq~(\ref{eq:adv-training}) indicates that adversarial training involves two processes: (1) generating perturbations that maximize the prediction loss and (2) updating model parameters that minimize the prediction loss. By alternating the above two processes iteratively, we can train a robust model against to adversarial attacks. Since there are two inputs, i.e., adjacency matrix ${\bf A}$ and attribute matrix ${\bf X}$, adversarial training can be done on them separately. To generate perturbations on the adjacency matrix, it is proposed to randomly drop edges during adversarial training~\cite{rl-s2v}. Though such simple strategy cannot lead to very significant improvement in classification accuracy (1\% increase), it shows some effectiveness with such cheap adversarial training. Furthermore, projected gradient descent is used to generate perturbations on the discrete input structure, instead of randomly dropping edges~\cite{xu2019topology-attack}. On the other hand, an adversarial training strategy with dynamic regularization is proposed to perturb the input features~\cite{feng2019graph-adversarial-training}. Specifically, it includes the divergence between the prediction of the target example and its connected examples into the objective of adversarial training, aiming to attack and reconstruct graph smoothness. Furthermore, batch virtual adversarial training~\cite{deng2019batch-adv-training} is proposed to promote the smoothness of GNNs and make GNNs more robust against adversarial perturbations. Several other variants of adversarial training on the input layer are introduced in~\cite{chen2019can-adv-training,dai2019adv-training-network-embedding,wang2019graphdefense-adv-training,dou2020robust}. 

The aforementioned adversarial training strategies face two main shortcomings: (1) they generate perturbations on ${\bf A}$ and ${\bf X}$ separately; and (2) it is not easy to perturb the graph structure due to its discreteness. To overcome the shortcomings, instead of generating perturbation on the input, a latent adversarial training method injects perturbations on the first hidden layer~\cite{jin2019latent-adversarial-training}:
\begin{equation}
\min _{\theta} \max _{\delta \in \mathcal{P}} \mathcal{L}_{\text{train}}\left(f_{\theta}(G;{\bf H}^{(1)}+\delta)\right),
\end{equation}
where ${\bf H}^{(1)}$ denotes the representation matrix of the first hidden layer and $\delta\in\mathcal{P}$ is some perturbation on ${\bf H}$.  It is noted that the hidden representation is continuous and it incorporates the information from both graph structure and node attributes.

\subsection{Adversarial Perturbation Detection}

To resist graph adversarial attacks during the test phase, there is one main strategy called adversary detection. These detection models protect the GNN models by exploring the intrinsic difference between adversarial edges/nodes and the clean edges/nodes~\cite{xu2018characterizing,ioannidis2019graphsac}. The work~\cite{xu2018characterizing} is the first work to propose detection approaches to find adversarial examples on graph data. It introduces four methods to distinguish adversarial edges or nodes from the clean ones including (1) link prediction (2) sub-graph link prediction (3) graph generation models and (4) outlier detection. These methods have shown some help to correctly detect adversarial perturbations. The work~\cite{ioannidis2019graphsac} introduces a method to randomly draw subsets of nodes, and relies on graph-aware criteria to judiciously filter out contaminated nodes and edges before employing a semi-supervised learning (SSL) module. The proposed model can be used to detect different anomaly generation models, as well as adversarial attacks.

\subsection{Certifiable Robustness}
Previously introduced adversarial training strategies are heuristic and only show experimental benefits. However, we still do not know whether there exist adversarial examples even when current attacks fail. Therefore, there are works~\cite{zugner2019certifiable-feature,bojchevski2019certifiable-graph,jia2020certified, bojchevski2020efficient, zugner2020certifiable3, tao2020adversarial, wang2020certified} considering to seriously reason the safety of graph neural networks which try to certify the GNN's robustness. As we know, GNN's prediction on one node $v_t$ always depends on its neighbor nodes. In~\cite{zugner2019certifiable-feature}, they ask the question: which nodes in a graph are safe under the risk of any admissible perturbations of its neighboring nodes' attributes. To answer this question, for each node $v$ and its corresponding label $y_v$, they try to exactly calculate an upper bound $U(v)$ of the maximized margin loss :
\begin{equation}\label{eq:certify}
U(v) \geq \max_{G'\in \mathcal{G}}\Big(\max _{i \neq y}  {\bf Z}_{v,i}\left( G^{\prime}\right)-{\bf Z}_{v,y}\left( G^{\prime}\right)\Big),
\end{equation}
where $\mathcal G$ denotes the set of all allowed graph perturbations (in~\cite{zugner2019certifiable-feature} only attribute perturbation). This upper bound $U$ is called the \textit{Certificate} of node $v$. During the certification, for $v$, if $U(v)\leq 0$, any perturbation cannot result the model to give a larger score to a wrong class $(i\neq y_v)$ than the class $y_v$, so there is no adversarial attack in $\mathcal{G}$ that can change the model's prediction. During the test phase, they calculate the certificate for all test nodes, thus they can know how many nodes in a graph is absolutely safe under attributes perturbation. Moreover, this certificate is trainable, directly minimizing the certificates will help more nodes become safe. However, the work~\cite{zugner2019certifiable-feature} only considers the perturbations on node attributes. Analyzing certifiable robustness from a different perspective, 
in~\cite{bojchevski2019certifiable-graph}, it deals with the case when the attacker only manipulates the graph structure. It derives the robustness certificates (similar to Eq.~(\ref{eq:certify})) as a linear function of personalized PageRank \cite{jeh2003scaling}, which makes the optimization tractable. In~\cite{zugner2020certifiable3}, it also tries to certify robustness of graph neural networks under graph structural perturbations. It successfully solves the certification problem using a jointly constrained bilinear programming method. In~\cite{wang2020certified}, it borrows the idea from \textit{randomized smoothing}~\cite{cohen2019certified} to achieve certifiable robustness for graphs under structural perturbation. Meanwhile, the sparsity of graph data during certification is considered in~\cite{bojchevski2020efficient}. It improves the efficiency and accuracy of the certifications against attacks on both graph features and structures. Immunization methods are introduced to improve graphs' certifiable robustness~\cite{tao2020adversarial}. Besides, there are also works studying certifiable robustness on GNN's other applications such as community detection~\cite{jia2020certified}.

\subsection{Graph Purification}
Both adversarial training or certifiable defense methods only target on resisting evasion attacks, which means that the attack happens during the test time. However, graph purification defense methods mainly focus on defending poisoning attacks. Since the poisoning attacks insert poisons into the training graph, purification methods aim to purify the poisoned graph and learn robust graph neural network models based on 
it. There are two approaches to realize graph purification: pre-processing~\cite{deep-insight-jaccard,entezari2020all-svd} and graph learning~\cite{jin2020graph,kipf2016variational}.

\subsubsection{Pre-processing}
Pre-processing methods first purify the perturbed graph data and then train the GNN model on the purified graph. In this way, the GNN model is trained on a clean graph. The work~\cite{deep-insight-jaccard} proposes a purification method based on two empirical observations of the attack methods: (1) Attackers usually prefer adding edges over removing edges or modifying features and (2) Attackers tend to connect dissimilar nodes. As a result, they propose a defense method by eliminating the edges whose two end nodes have small Jaccard Similarity~\cite{said2010social}. Because these two nodes are different and it is not likely they are connected in reality,  the edge between them may be adversarial. The experimental results demonstrate the effectiveness and efficiency of the proposed defense method. However, this method can only work when the node features are available. In~\cite{entezari2020all-svd}, it is observed that Nettack~\cite{nettack} generates the perturbations which mainly changes the small singular values of the graph adjacency matrix. Thus it proposes to purify the perturbed adjacency matrix by using truncated SVD to get its low-rank approximation. it further shows that only keeping the top $10$ singular values of the adjacency matrix is able to defend Nettack and improve the performance of GNNs.

\subsubsection{Graph Learning}
Pre-procesing might not be an ideal choice for purifying the graph, as it is independent of GNN's training process and could mistakenly remove normal edges. An alternative purification strategy is graph learning, which targets at removing adversarial patterns and obtaining clean graph structure. Traditional graph learning methods~\cite{franceschi2019learning,jiang2019semi} do not directly deal with adversarial attacks. Hence, to make GNNs more robust, it is of great importance to leverage the characteristics of adversarial attacks to guide the graph learning process. In~\cite{jin2020graph}, it has demonstrated that adversarial attacks can essentially violate some important graph properties, i.e, low-rank, sparsity and feature smoothness. Then it explores these graph properties to design robust graph neural networks, Pro-GNN, which jointly learns clean graph structure and robust GNN parameters. Specifically, Pro-GNN alternatively updates graph structure by preserving the aforementioned properties and trains GNN parameters on the updated graph structure. In~\cite{zhang2020defensevgae}, variational graph autoencoder~\cite{kipf2016variational} is employed to reconstruct graph structure, which can also reduce the effects of adversarial perturbations.

\subsection{Attention Mechanism}
Different from the purification methods which try to exclude adversarial perturbations, attention-based defense methods aim to train a robust GNN model by penalizing model's weights on adversarial edges or nodes~\cite{rgcn,pa-gnn,zhang2020gnnguard}. 
Basically, these methods learn an attention mechanism to distinguish adversarial edges and nodes from the clean ones, and then make the adversarial perturbations contribute less to the aggregation process of the GNN training. The work~\cite{rgcn} assumes that adversarial nodes may have high prediction uncertainty, since adversary tends to connect the node with nodes from other communities. In order to penalize the influence from these uncertain nodes, they propose a defense method named RGCN to model the $l$-th layer hidden representation $\boldsymbol{h}_i^{(l)}$ of nodes as Gaussian distribution with mean value $\boldsymbol{\mu}_{\mathrm{i}}^{(l)}$ and variance $\boldsymbol{\sigma}_{i}^{(l)}$, 
\begin{equation}
\boldsymbol{h}_{i}^{(l)} \sim N\left(\boldsymbol{\mu}_{\mathrm{i}}^{(l)}, \operatorname{diag}\left(\boldsymbol{\sigma}_{i}^{(l)}\right)\right),
\end{equation}
where the uncertainty can be reflected in the variance $\boldsymbol{\sigma}_{i}^{(l)}$. When aggregating the information from neighbor nodes, it applies an attention mechanism to penalize the nodes with high variance,
\begin{equation}
\boldsymbol{\alpha}_{i}^{(l)}=\exp \left(-\gamma \boldsymbol{\sigma}_{i}^{(l)}\right),
\end{equation}
where $\boldsymbol{\alpha}^{(l)}_i$ is the attention score assigned to node $i$ and $\gamma$
is a hyper-parameter. Furthermore, it is verified that the attacked nodes do have higher variances than normal nodes and the proposed attention mechanism does help mitigate the impact brought by adversarial attacks. 
From a different perspective, GNNGuard is proposed to employ the theory of network homophily~\cite{mcpherson2001birds} to assign higher scores to edges connecting similar nodes while pruning
edges between unrelated node~\cite{zhang2020gnnguard}.

In~\cite{pa-gnn}, it suggests that to improve the robustness of one target GNN model, it is beneficial to include the information from other clean graphs, which share the similar topological distributions and node attributes with the target graph. For example, Facebook and Twitter have social network graph data that share similar domains; Yelp and Foursquare have similar co-review graph data. Thus, it first generates adversarial edges $E_P$ on the clean graphs, which serve as the supervision of known perturbation. With this supervision knowledge, it further designs the following loss function to reduce the attention scores of adversarial edges:
\begin{equation}
\mathcal{L}_{dist}=-\min \left(\eta, \underset{e_{i j} \in {E} \backslash {E_P}}{\mathbb{E}} \boldsymbol{\alpha}_{i j}^{(l)}-\underset{e_{ij} \in {E_P}}{\mathbb{E}} \boldsymbol{\alpha}_{i j}^{(l)}\right),
\end{equation}
where $\mathbb{E}$ denotes the expectation, $E\backslash{E_P}$ represents normal edges in the graph, $\boldsymbol{\alpha}^{(l)}_{ij}$ is the attention score assigned to edge $e_{ij}$ and $\eta$ is a hyper-parameter controlling the margin between the expectation of two distributions. It then adopts meta-optimization to train a model initialization and fine-tunes it on the target poisoned graph to get a robust GNN model.


\section{A Repository for Graph Attacks and Defenses}
In this section, we give a brief introduction to the repository we have developed for adversarial attacks and defenses, DeepRobust\footnote{{https://github.com/DSE-MSU/DeepRobust}}~\cite{li2020deeprobust}.
To facilitate the research on adversarial attacks and defenses, DeepRobust includes the majority of representative attack and defense algorithms for both graph data and image data. The repository can enable researchers to deepen our understandings on attacks and defenses via empirical studies. Specifically, for graph adversarial learning, DeepRobust provides 7 datasets: four citation graphs including Cora~\cite{sen2008collective}, Cora-ML~\cite{bojchevski2017deep}, Citeseer~\cite{bojchevski2017deep} and Pubmed~\cite{sen2008collective}, one co-author graph ACM~\cite{wang2019heterogeneous}, one blog 
graph Polblogs~\cite{adamic2005political} and one social graph BlogCatalog~\cite{huang2017label}. On top of that, DeepRobust covers the following attacks and defenses: (1) 5 targeted attack algorithms, i.e., FGA~\cite{chen2018fast-gradient-network-embedding}, Nettack~\cite{nettack}, RL-S2V~\cite{rl-s2v}, integrated gradient attack~\cite{deep-insight-jaccard} and random attack~\cite{nettack}; (2) 3 untargeted algorithms, i.e., Metattack~\cite{metattack}, Topology attack~\cite{xu2019topology-attack} and DICE~\cite{Waniek_2018}; (3) one victim model, i.e. GCN~\cite{kipf2016semi}; and (4) 5 defense algorithms, i.e., adversarial training~\cite{goodfellow2014explaining}, GCN-Jaccard~\cite{deep-insight-jaccard}, GCN-SVD~\cite{entezari2020all-svd}, RGCN~\cite{rgcn} and Pro-GNN~\cite{jin2020graph}. 

DeepRobust is an easy-to-use platform for researchers who are working on adversarial attacks and defenses. With DeepRobust, users can generate graph adversarial attacks by training an attack model through the attacking APIs or loading the pre-attacked graphs provided by the repository. Robust models can be trained on the perturbed graph with the defense APIs.  Once we obtain the perturbed graph and the trained defense model, they can be fed into the evaluation API to compete against each other, and the model performance under the corresponding attack will be reported. In the next section, we use DeepRobust for empirical studies on graph adversarial attacks.

\section{Empirical Studies}
With the help of DeepRobust, we are able to conduct empirical studies on graph adversarial attacks and discover their important patterns. Next we first introduce the experimental settings and then present the empirical results and findings. 

\subsection{Experimental Setup}

Different attack and defense methods have been designed under different settings. We perform the experiments with one of the most popular settings -- the untargeted poisoning setting. Correspondingly we choose representative attack and defense methods that have been designed for this setting. Three representative attack methods are adopted to generate perturbations including DICE~\cite{Waniek_2018}, Metattack~\cite{metattack} and Topology attack~\cite{xu2019topology-attack}. It is noted that DICE is a white-box attack which randomly connects nodes with different labels or drops edges between nodes sharing the same label. To evaluate the performance of different defense methods under adversarial attacks, we compare the robustness of the natural trained GCN~\cite{kipf2016semi} and four defense methods on those attacked graphs, i.e., GCN~\cite{kipf2016semi}, GCN-Jaccard~\cite{deep-insight-jaccard}, GCN-SVD~\cite{entezari2020all-svd}, RGCN~\cite{rgcn} and GAT~\cite{velivckovic2017graph}. Following~\cite{metattack}, we use three datasets: Cora, Citeseer and Polblogs. For each dataset, we randomly choose 10\% of nodes for training, 10\% of nodes for validation and the remaining 80\% for test. We repeat each experiment for 5 times and report the average performance. On Cora and Citeseer datasets, the most destructive variant CE-min-max~\cite{xu2019topology-attack} is adopted to implement Topology attack. But CE-min-max cannot converge on Polblogs dataset, we adopt another variant called CE-PGD~\cite{xu2019topology-attack} on this dataset.

\subsection{Analysis on Attacked Graph}

One way to understand the behaviors of attacking methods is to compare the properties of the clean graph and the attacked graph. In this subsection, we perform this analysis from both global and local perspectives.
\vskip 0.5em
\noindent{}\textbf{Global Measure.} We have collected five global properties from both clean graphs and perturbed graphs generated by the three attacks on the three datasets. These properties include the number of added edges, the number of deleted edges, the number of edges, the rank of the adjacent matrix, and clustering coefficient. We only show the results of Metattack in Table~\ref{tab:pro-meta}. Results for Topology attack and DICE can be found in Appendix A.1. Note that we vary the perturbation rates from $0$ to $25\%$ with a step of $5\%$ and $0\%$ perturbation denotes the original clean graph. It can be observed from the table: 
\begin{itemize}
    \item  Attackers favor adding edges over deleting edges.
    \item Attacks are likely to  increase the rank of the adjacency matrix. 
    \item Attacks are likely to reduce the connectivity of a graph. The clustering coefficients of a perturbed graph decrease with the increase of the perturbation rate.
\end{itemize}

\begin{table}[h]
{
\centering
\caption{Properties of attacked graphs under Metattack. Note that $r$ denotes perturbation rate and $0\%$ perturbation indicates the original clean graph. }
\resizebox{0.47\textwidth}{!}{
\begin{tabular}{lllllll}
\toprule
Dataset                   & $r$(\%) & edge+  & edge-  & edges   & ranks & \begin{tabular}[c]{@{}l@{}}clustering \\ coefficients\end{tabular}  \\ \hline
\multirow{6}{*}{Cora}     & 0    & 0      & 0      & 5069    & 2192  & 0.2376                       \\   
                          & 5    & 226  & 27   & 5268  & 2263  & 0.2228               \\   
                          & 10   & 408  & 98   & 5380  & 2278  & 0.2132                \\   
                          & 15   & 604  & 156  & 5518  & 2300  & 0.2071                \\   
                          & 20   & 788  & 245  & 5633  & 2305  & 0.1983               \\   
                          & 25   & 981  & 287  & 5763  & 2321  & 0.1943               \\ \hline
\multirow{6}{*}{Citeseer} & 0    & 0      & 0      & 3668    & 1778  & 0.1711                      \\   
                          & 5    & 181  & 2    & 3847  & 1850  & 0.1616                \\   
                          & 1    & 341  & 25   & 3985  & 1874  & 0.1565              \\   
                          & 15   & 485  & 65   & 4089  & 1890  & 0.1523            \\   
                          & 20   & 614  & 119  & 4164 & 1902  & 0.1483                \\   
                          & 25   & 743  & 174  & 4236  & 1888  & 0.1467                 \\ \hline
\multirow{6}{*}{Polblogs} & 0    & 0      & 0      & 16714   & 1060  & 0.3203                      \\   
                          & 5    & 732  & 103  & 17343 & 1133  & 0.2719            \\   
                          & 10   & 1347 & 324  & 17737 & 1170  & 0.2825              \\   
                          & 15   & 1915 & 592  & 18038 & 1193  & 0.2851                 \\   
                          & 20   & 2304   & 1038   & 17980   & 1193  & 0.2877              \\   
                          & 25   & 2500 & 1678 & 17536 & 1197  & 0.2723               \\ 
\bottomrule
\label{tab:pro-meta}
\end{tabular}
}
}
\end{table}

\begin{figure}[h]
	\begin{center}
	  \subfigure[Cora] {\includegraphics[scale=0.25]{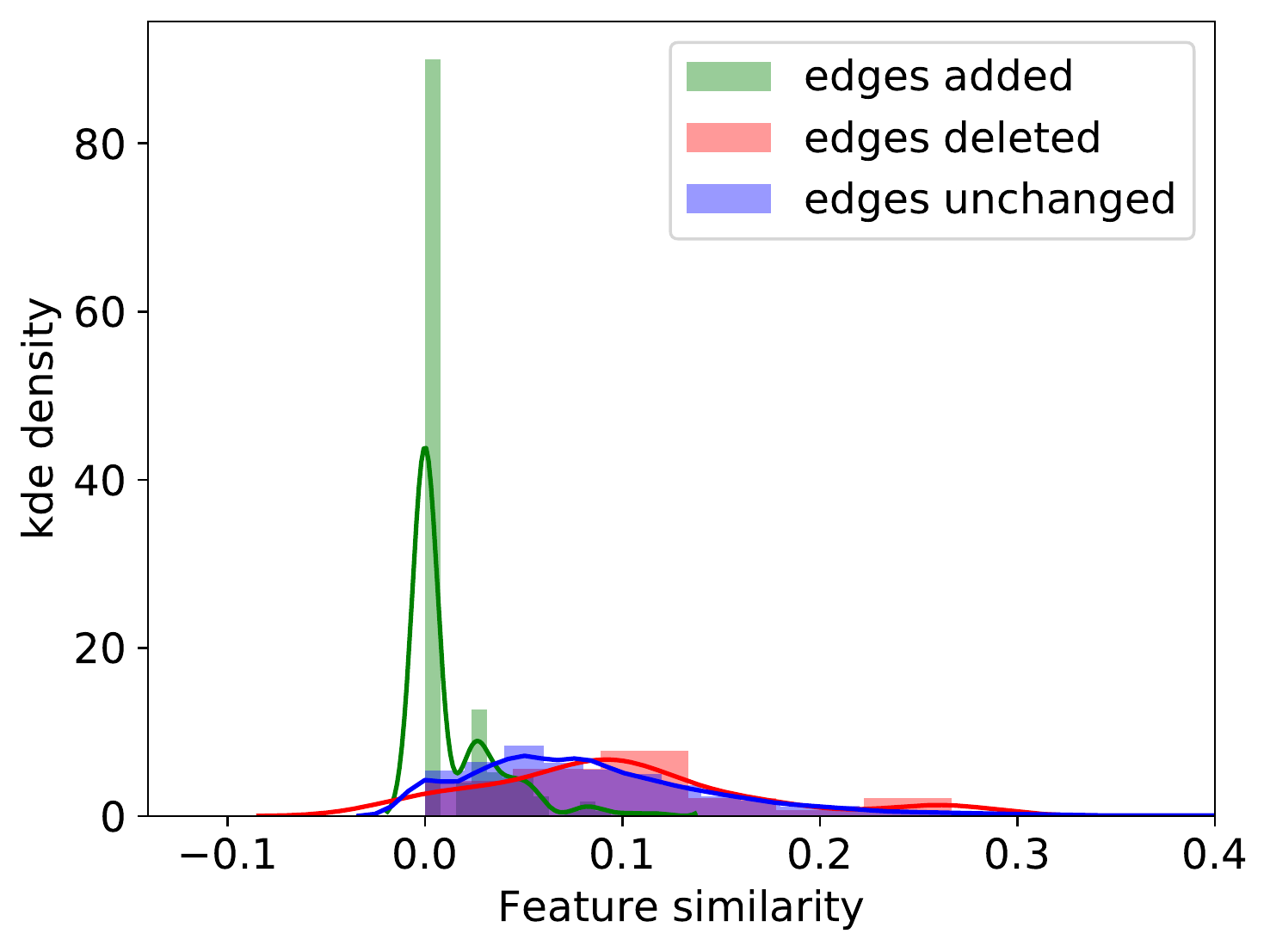}} 
	  \subfigure[Citeseer]{\includegraphics[scale=0.25]{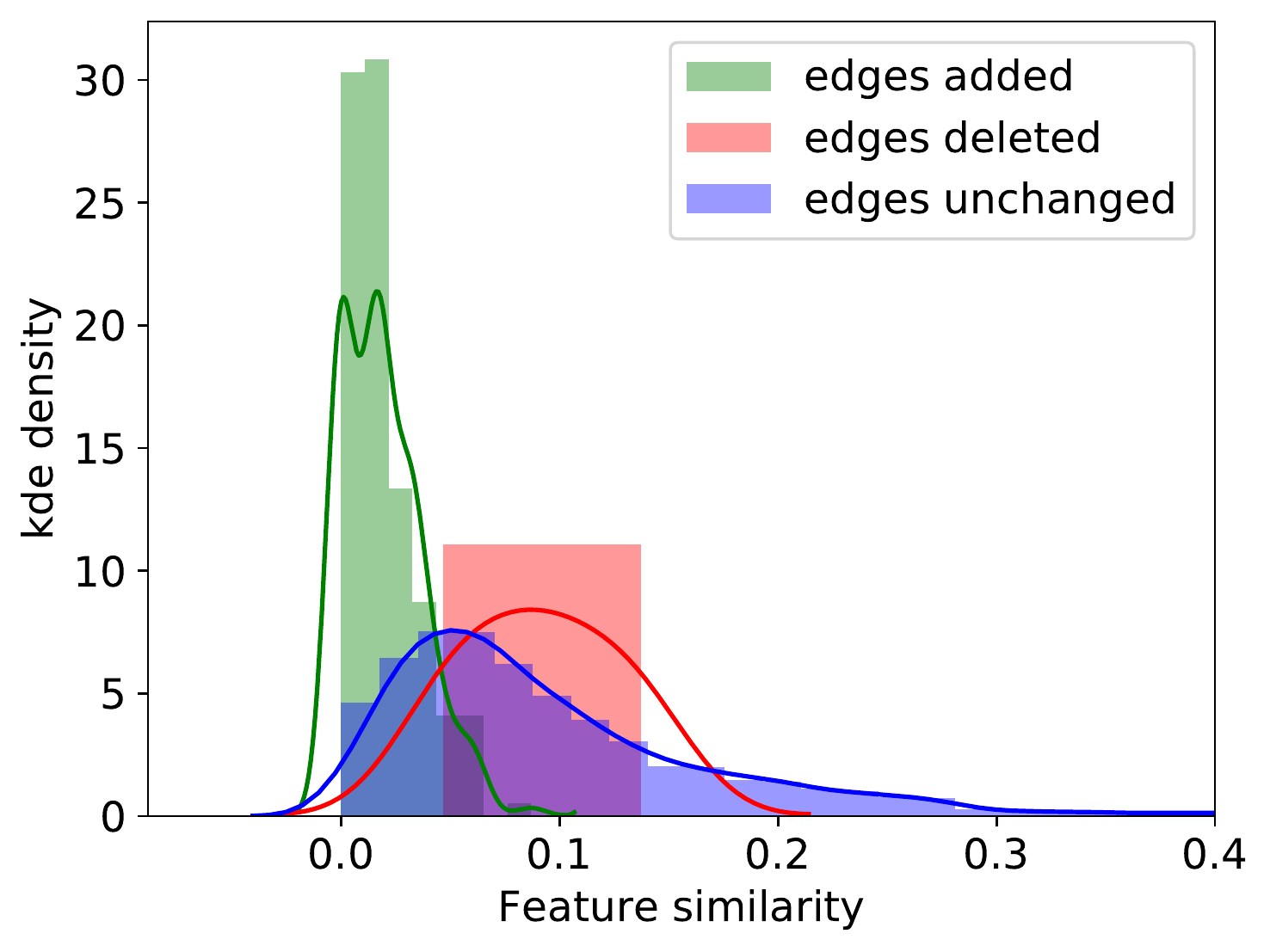}}
	\end{center}
\vskip -1em
\caption{Node feature similarity for Metattack.}
\label{fig:featuresimilarity}
\end{figure}

\begin{figure*}[h]
	\begin{center}
	  \subfigure[Cora] {\includegraphics[scale=0.33]{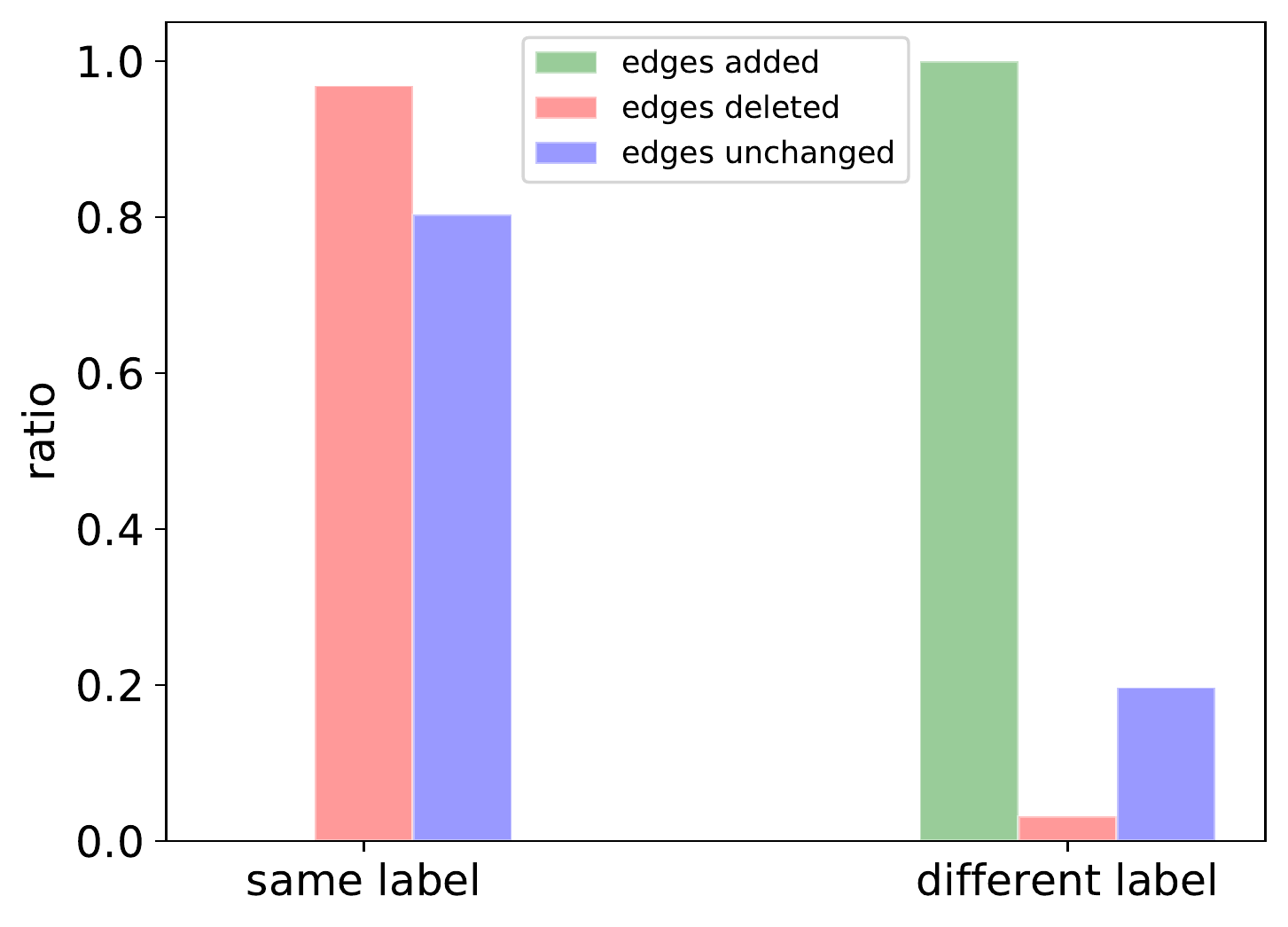}} 
	  \subfigure[Citeseer]{\includegraphics[scale=0.33]{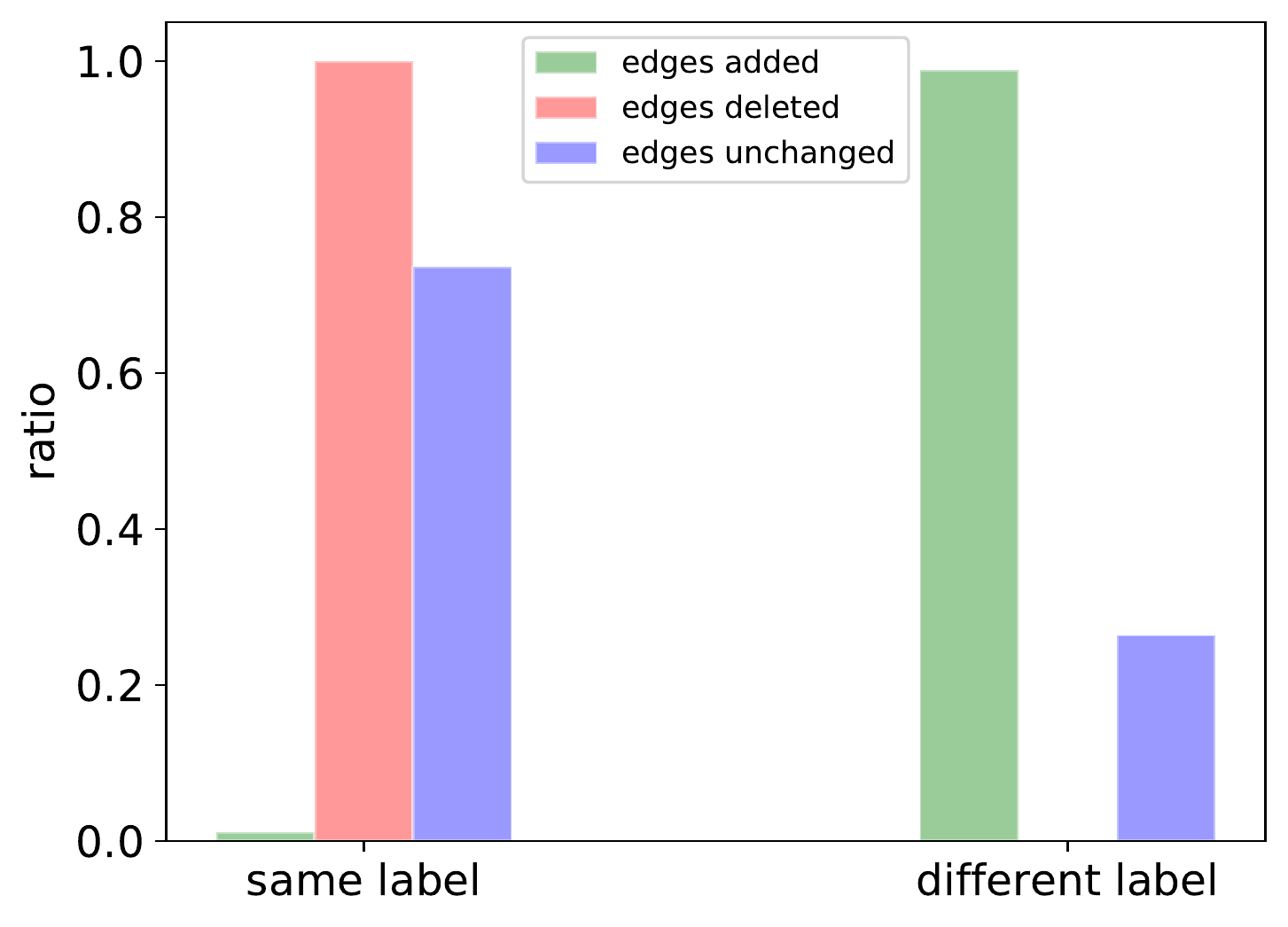}}
	  \subfigure[Polblogs] {\includegraphics[scale=0.33]{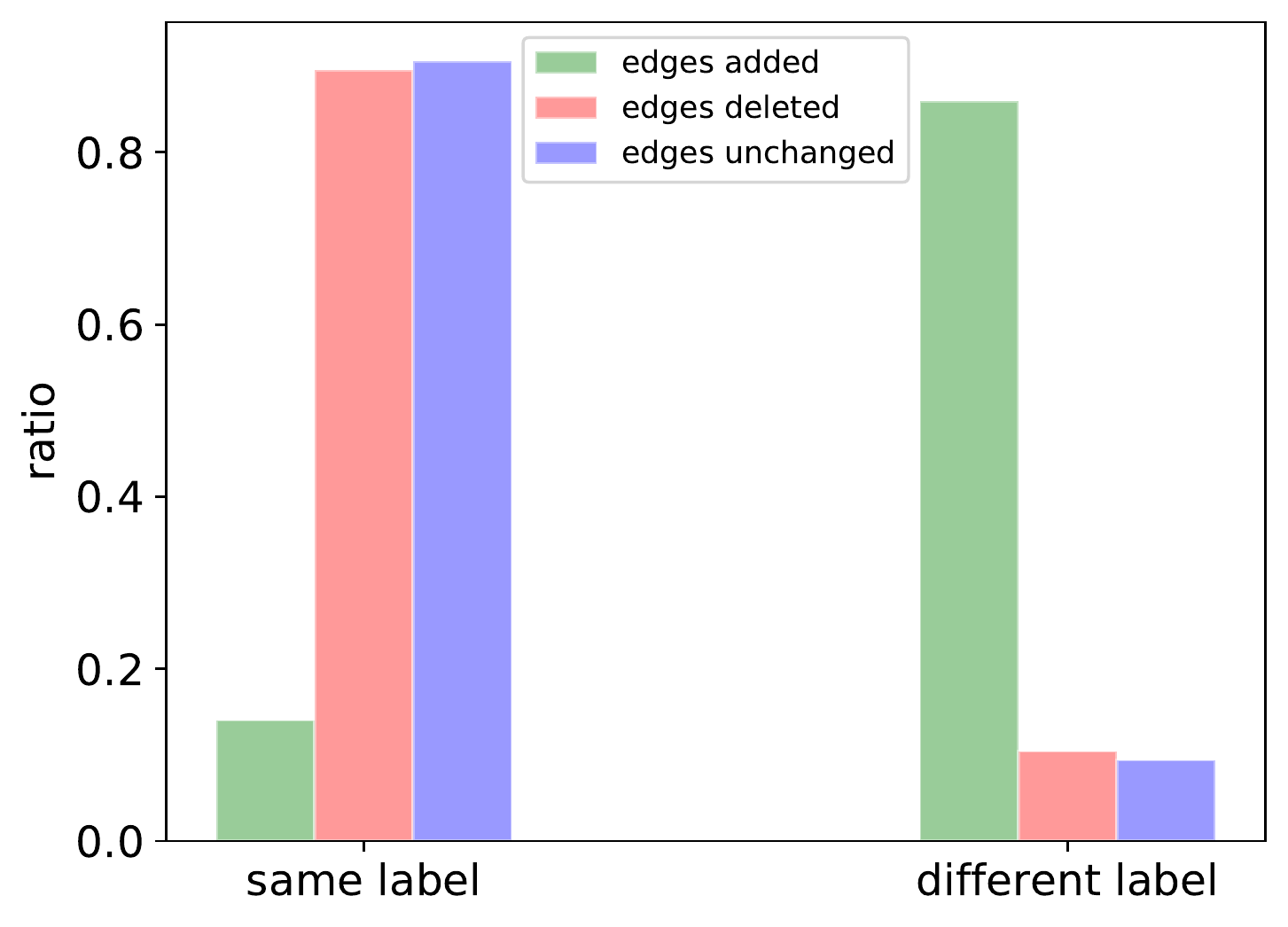}} 
	\end{center}
\vskip -1em
\caption{Label equality for Metattack.}
\label{fig:labequality}
\end{figure*}

\vskip 0.5em
\noindent{}\textbf{Local Measure.} We have also studied two local properties including the feature similarity and label equality between two nodes connected by three kinds of edges: the newly added edges, the deleted edges and the normal edges which have not been changed by the attack methods. Since features are binary in our datasets, we use jaccard similarity as the measure for feature similarity. For label equality, we report the ratio if two nodes share the same label or have different labels. The feature similarity and label equality results are demonstrated in Figures~\ref{fig:featuresimilarity} and~\ref{fig:labequality}, respectively. We show the results for Metattack with $5\%$ perturbations. Results for Topology attack and DICE can be found in Appendix A.2. Note that we do not have feature similarity results on Polblogs since this dataset does not have node features. We can make the following observations from the figures. 
\begin{itemize}
    \item {Attackers tend to connect nodes with dissimilar features and different labels. As shown in Figure~\ref{fig:featuresimilarity} and Figure~\ref{fig:labequality}, most of the added edges connect nodes with very dissimilar features and different labels.}
    \item {Attackers tend to remove edges from nodes which share similar features and same label. As shown in Figure~\ref{fig:featuresimilarity} and Figure~\ref{fig:labequality}, most of the deleted edges originally connect nodes sharing similar features and same labels.}
\end{itemize}

%

\subsection{Attack and Defense Performance}

In this subsection, we study how the attack methods perform and whether the defense methods can help resist to attacks. Similarly, we vary the perturbation rates from $0$ to $25\%$ with a step of $5\%$. The results are demonstrated in Table~\ref{tab:meta}. We show the performance for Metattack. Results for Topology attack and DICE are shown in Appendix A.3. Note that we do not have the performance for Jaccard defense model in Polblogs since this mode requires node features and Polblogs does not provide node features. According to the results, we have the following observations:
\begin{itemize}
   \item With the increase of the perturbations, the performance of GCN dramatically deceases. This result suggests that Metattack can lead to a significant reduce of accuracy on the GCN model. 
   \item When the perturbations are small, we observe small performance reduction for defense methods which suggests their effectiveness. However, when the graphs are heavily poisoned, their performance also reduces significantly which indicates that efforts are needed to defend heavily poisoning attacks.  
\end{itemize}
\begin{table}[htbp]

    \caption{Performance (accuracy) under Metattack}
    \begin{center}
    \begin{threeparttable}
    \centering
\resizebox{0.47\textwidth}{!}{
    \begin{tabular}{@{}ccccccccc@{}}
    \toprule
    Dataset& $r$ (\%) & 0 & 5 & 10 & 15 & 20 & 25 \\
    \midrule
    \multirow{5}{*}{Cora }& GCN & 83.10 & 76.69 & 65.58 & 54.88 & 48.66 & 38.44\\

    & Jaccard\footnote[1]{} & 82.39 & 81.02 & 77.28 & 72.74 & 69.16 & 64.56 \\

    & SVD\footnote[2]{} & 77.97 & 75.67 & 70.51 & 64.34 & 55.89 & 45.92\\

    & RGCN & 84.81 & 81.32 & 72.12 & 60.25 & 49.75 & 37.76\\

    &GAT & 81.69 & 74.75 & 61.69 & 52.56 & 45.30 & 38.52\\

    \midrule

    \multirow{5}{*}{Citeseer}& GCN & 74.53 & 72.59 & 63.96 & 61.66 & 50.58 & 44.32 \\
    
    & Jaccard\footnote[1]{} & 74.82 & 73.60 & 73.50 & 72.80 & 72.97 & 72.53\\
    
    & SVD\footnote[2]{} & 70.32 & 71.30 & 67.58 & 63.86 & 56.91 & 45.28\\
    
    & RGCN & 74.41 & 72.68 & 71.15 & 69.38 & 67.93 & 67.24\\
    
    & GAT & 74.23 & 72.01 & 67.12 & 57.70 & 47.97 & 38.70\\
    \midrule

    \multirow{4}{*}{Polblogs}& GCN & 95.80 & 73.93 & 72.07 & 67.69 & 62.29 & 52.97 \\

    & SVD\footnote[2]{} & 94.99 & 82.64 & 71.27 & 66.09 & 61.37 & 52.82\\

    & RGCN & 95.60 & 72.01 & 67.12 & 57.70 & 47.97 & 38.70\\
    
    & GAT & 95.40 & 84.83 & 77.03 & 69.94 & 53.62 & 53.76\\
    \bottomrule
    \end{tabular}}
    \begin{tablenotes}
    \small
    \item[1] Jaccard: GCN-Jaccard defense model. 
    \item[2] SVD: GCN-SVD defense model.
    \end{tablenotes}
\end{threeparttable}
\end{center}
\label{tab:meta}
\vskip -1em
\end{table}

\section{Conclusion \& Future Directions}
In this survey, we give a comprehensive overview of an emerging research field, adversarial attacks and defenses on graph data. We investigate the taxonomy of graph adversarial attacks, and review representative adversarial attacks and the corresponding countermeasures. Furthermore, we conduct empirical study to show how different defense methods behave under different attacks, as well as the changes in important graph properties by the attacks. Via this comprehensive study, we have gained deep understandings on this area that enables us to discuss some promising research directions. 
\begin{itemize}
    \item \textbf{Imperceptible perturbation measure.} Different from image data, humans cannot easily tell whether a perturbation on graph is imperceptible or not. The $\ell_0$ norm constraint on perturbation is definitely not enough. Currently only very few existing work study this problem, thus finding concise perturbation evaluation measure is of great urgency.  
    \item \textbf{Different graph data.} Existing works mainly focus on static graphs with node attributes. Complex graphs such as graphs with edge attributes and dynamic graphs are not well-studied yet. 
    \item \textbf{Existence and transferability of graph adversarial examples.} There are only a few works discussing about the existence and transferability of graph adversarial examples. Studying this topic is important for us to understand our graph learning algorithm, thus helping us build robust models.
    \item \textbf{Scalability.} The high complexity of attack methods has hindered their use in practical applications. However, there are only few works developing efficient attack methods in terms of time complexity. Furthermore, given that most of attack algorithms are gradient-based, how to reduce their memory complexity also remains a challenge.
\end{itemize}

\section*{Acknowledgments}
This research is supported by the National Science Foundation (NSF) under grant number CNS1815636, IIS1928278, IIS1714741, IIS1845081, IIS1907704, and IIS1955285.

\bibliographystyle{abbrv}
\bibliography{main} 

\vskip 3em
\appendix

\section{Additional Results}

\subsection{Global Measures for Topology Attack and DICE}
The global measures for Topology attack are shown in Table~\ref{tab:pro-top}; the global measures for DICE are shown in Table~\ref{tab:pro-dice}.

\begin{table}[h]
\centering

\caption{Properties of attacked graphs under Topology attack.}
\resizebox{0.47\textwidth}{!}{\begin{tabular}{llllllll}

\toprule
Dataset                   & $r$(\%)   & edges+ & edges- & edges   & ranks & \begin{tabular}[c]{@{}l@{}}clustering \\ coefficients\end{tabular}  \\ \hline
\multirow{6}{*}{Cora}     & 0    & 0      & 0      & 5069    & 2192  & 0.2376                    \\    
                          & 5 & 255    & 0      & 5324    & 2292  & 0.2308              \\    
                          & 10  & 508    & 0      & 5577    & 2369  & 0.2185              \\    
                          & 15 & 762    & 0      & 5831    & 2417  & 0.2029            \\    
                          & 20  & 1015   & 0      & 6084    & 2442  & 0.1875               \\    
                          & 25 & 1269   & 0      & 6338    & 2456  & 0.1736           \\ \hline
\multirow{6}{*}{Citeseer} & 0    & 0      & 0      & 3668    & 1778  & 0.1711                       \\    
                          & 5 & 185    & 0      & 3853    & 1914  & 0.1666               \\    
                          & 10  & 368    & 0      & 4036    & 2003  & 0.1568             \\    
                          & 15 & 552    & 0      & 4220    & 2058  & 0.1429               \\    
                          & 20  & 735    & 0      & 4403    & 2077  & 0.1306           \\    
                          & 25 & 918    & 0      & 4586    & 2087  & 0.1188             \\ \hline
\multirow{6}{*}{Polblogs} & 0    & 0      & 0      & 16714   & 1060  & 0.3203                    \\    
                          & 5 & 716    & 96   & 17334 & 1213  & 0.2659              \\    
                          & 10  & 1532 & 128  & 18118   & 1220  & 0.2513             \\    
                          & 15 & 2320 & 146  & 18887 & 1221  & 0.2408            \\    
                          & 20  & 3149 & 155    & 19708 & 1221  & 0.2317          \\    
                          & 25 & 3958 & 163  & 20509 & 1221  & 0.2238            \\ 
\bottomrule
\label{tab:pro-top}
\end{tabular}
}
\end{table}

\begin{table}[h]
\centering
\caption{Properties of attacked graphs under DICE attack.}
\resizebox{0.47\textwidth}{!}{\begin{tabular}{llllllll}
\toprule
Dataset                   & $r$(\%)    & edge+  & edge-  & edges   & ranks & \begin{tabular}[c]{@{}l@{}}clustering \\ coefficients\end{tabular}  \\ \hline
\multirow{6}{*}{Cora}     & 0    & 0      & 0      & 5069    & 2192  & 0.2376                  \\    
                          & 5 & 125  & 128  & 5066  & 2208  & 0.2165                \\    
                          & 10  & 251  & 255  & 5065  & 2241  & 0.1963             \\    
                          & 15 & 377  & 383  & 5063  & 2256  & 0.1768               \\    
                          & 20  & 504  & 509  & 5063  & 2262  & 0.1588                   \\    
                          & 25 & 625  & 642  & 5053  & 2271  & 0.1436                   \\ \hline
\multirow{6}{*}{Citeseer} & 0    & 0      & 0      & 3668    & 1798  & 0.1711               \\    
                          & 5 & 91     & 92     & 3667    & 1829  & 0.1574              \\    
                          & 10  & 183 & 183  & 3668  & 1843  & 0.1404                 \\    
                          & 15 & 276  & 274  & 3670  & 1858  & 0.1279               \\    
                          & 20  & 368 & 365  & 3672  & 1872  & 0.1170                    \\    
                          & 25 & 462    & 455    & 36755   & 1871  & 0.1068                  \\ \hline
\multirow{6}{*}{Polblogs} & 0    & 0      & 0      & 16714   & 1060  & 0.3203   \\    
                          & 5 & 420  & 415  & 16719 & 1151  & 0.2742              \\    
                          & 10  & 846  & 825  & 16736 & 1191  & 0.2341               \\    
                          & 15 & 1273 & 1234 & 16752 & 1206  & 0.2077              \\    
                          & 20  & 1690 & 1652 & 16752 & 1214  & 0.1862             \\    
                          & 25 & 2114 & 2064 & 16765 & 1216  & 0.1675                 \\ 
\bottomrule
\label{tab:pro-dice}
\end{tabular}
}

\end{table}

\subsection{Local Measures for Topology Attack and DICE}
\begin{figure}[htbp]
	\begin{center}
	  \subfigure[Cora] {\includegraphics[scale=0.25]{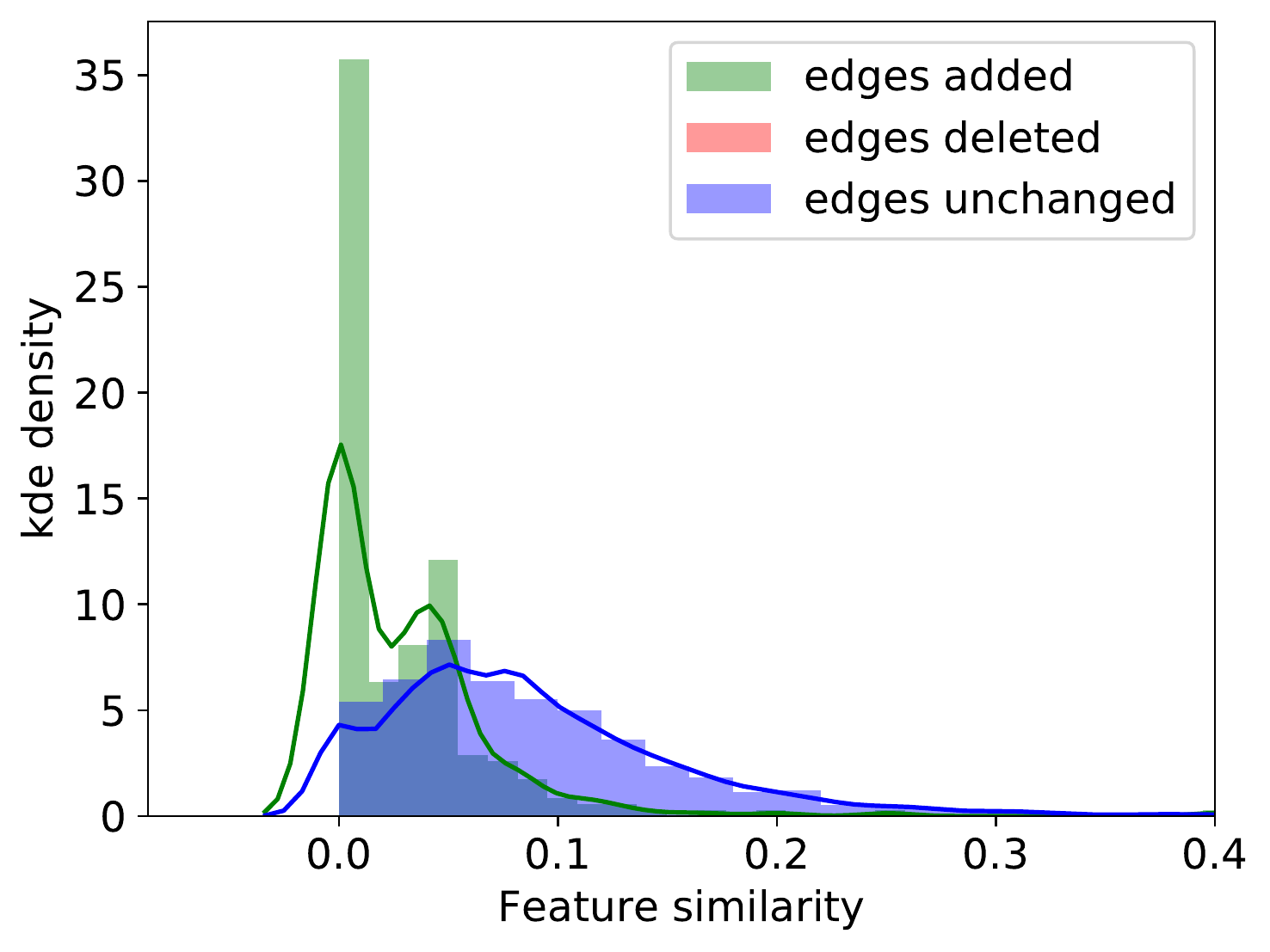}} 
	  \subfigure[Citeseer]{\includegraphics[scale=0.25]{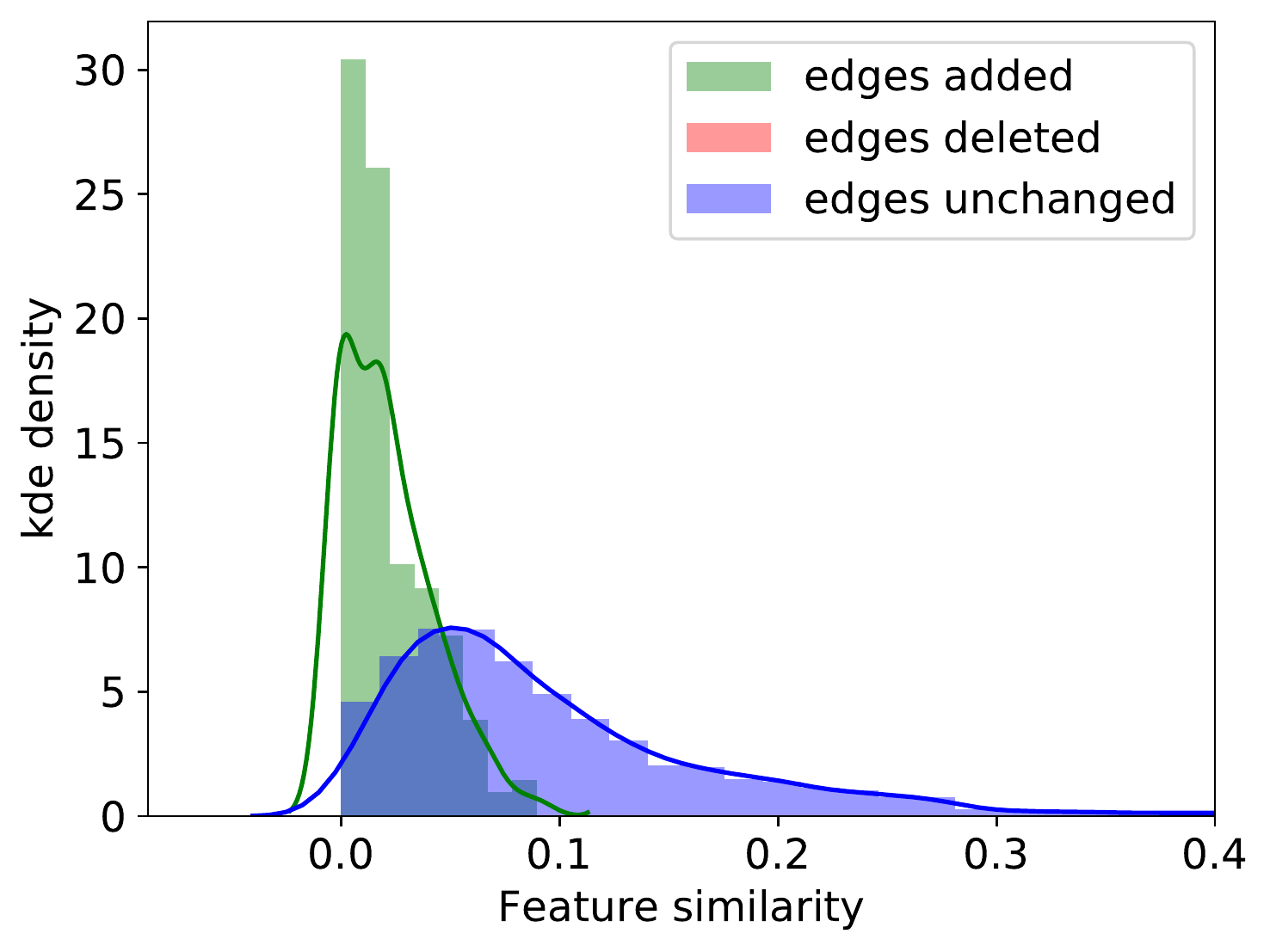}}
	\end{center}
\caption{Node feature similarity for Topology attack.}

\label{fig:featuresimilarity-top}
\end{figure}

\begin{figure}[htbp]
	\begin{center}
	  \subfigure[Cora] {\includegraphics[scale=0.25]{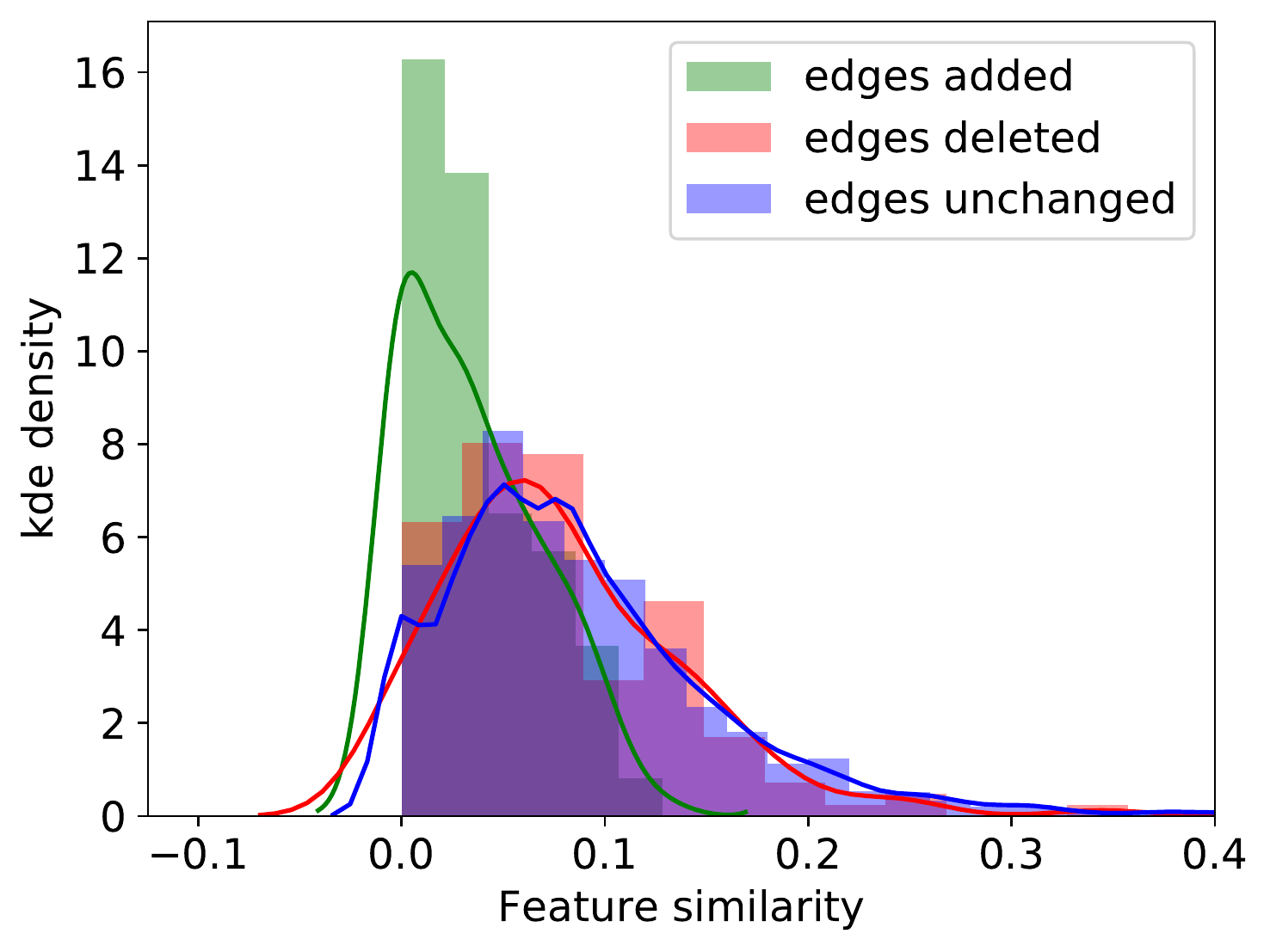}} 
	  \subfigure[Citeseer]{\includegraphics[scale=0.25]{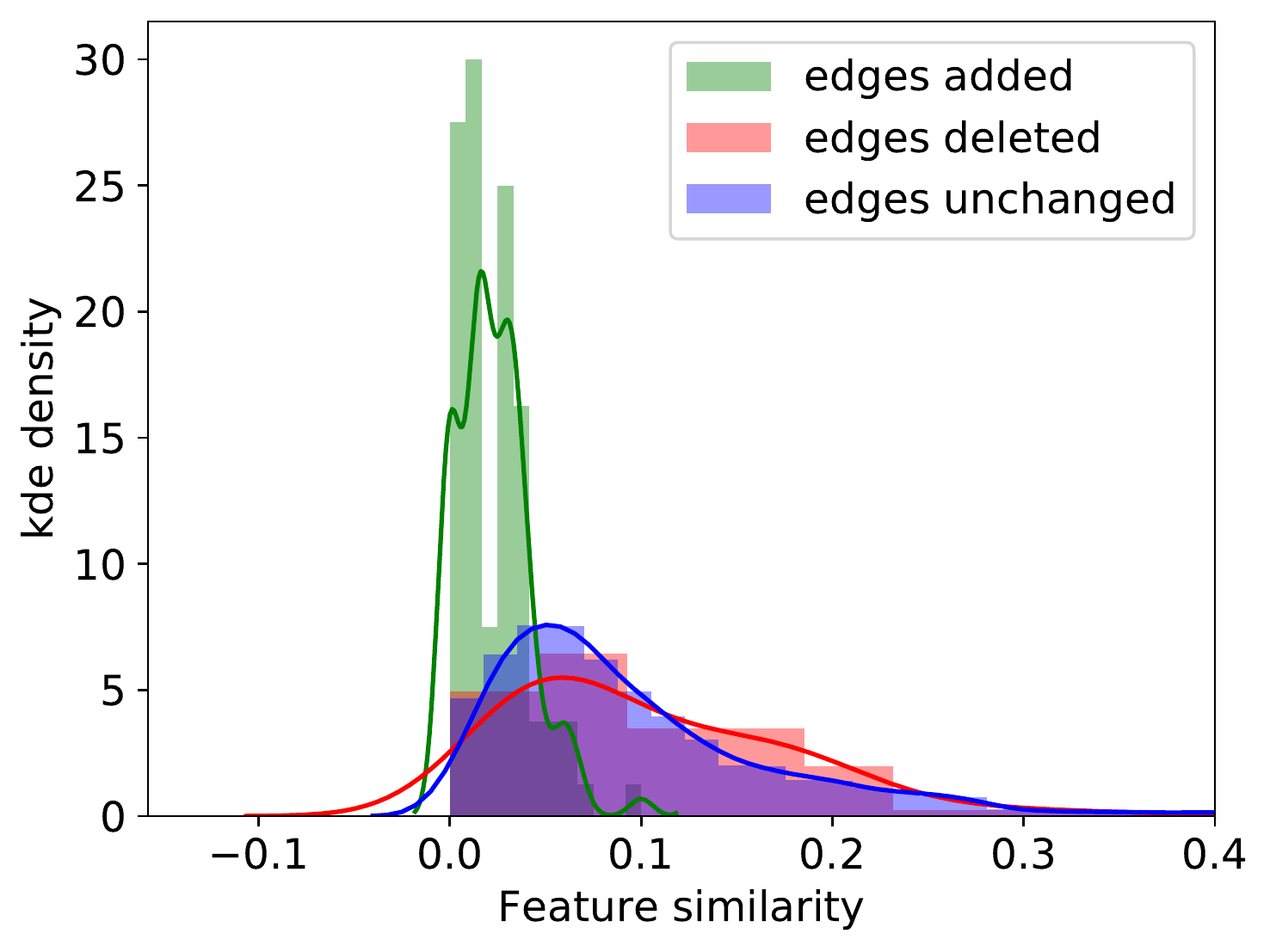}}		
	\end{center}
\caption{Node feature similarity for DICE attack.}
\label{fig:featuresimilarity-dice}
\end{figure}

\begin{figure*}[htbp]
	\begin{center}
	  \subfigure[Cora] {\includegraphics[scale=0.33]{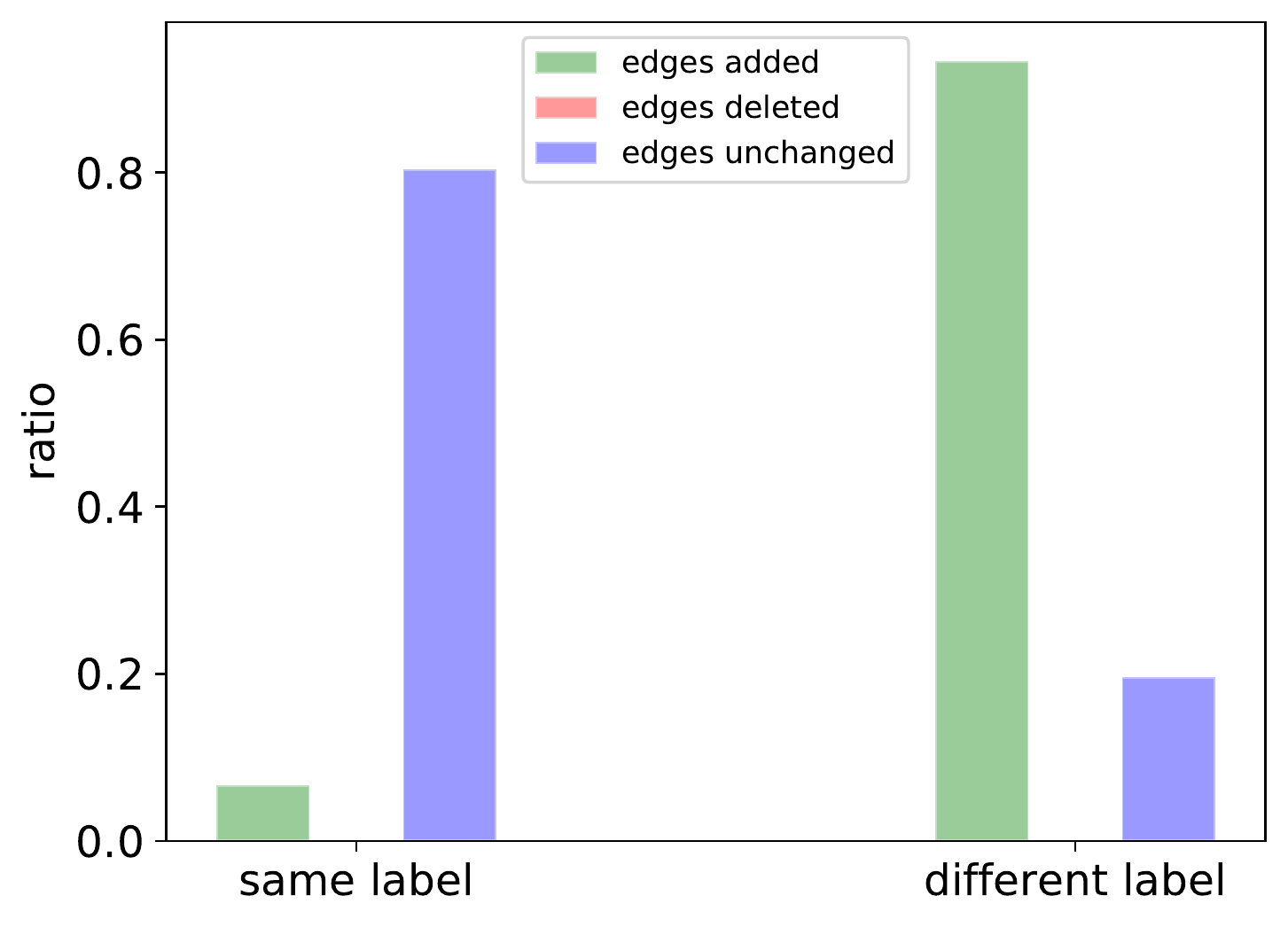}} 
	  \subfigure[Citeseer]{\includegraphics[scale=0.33]{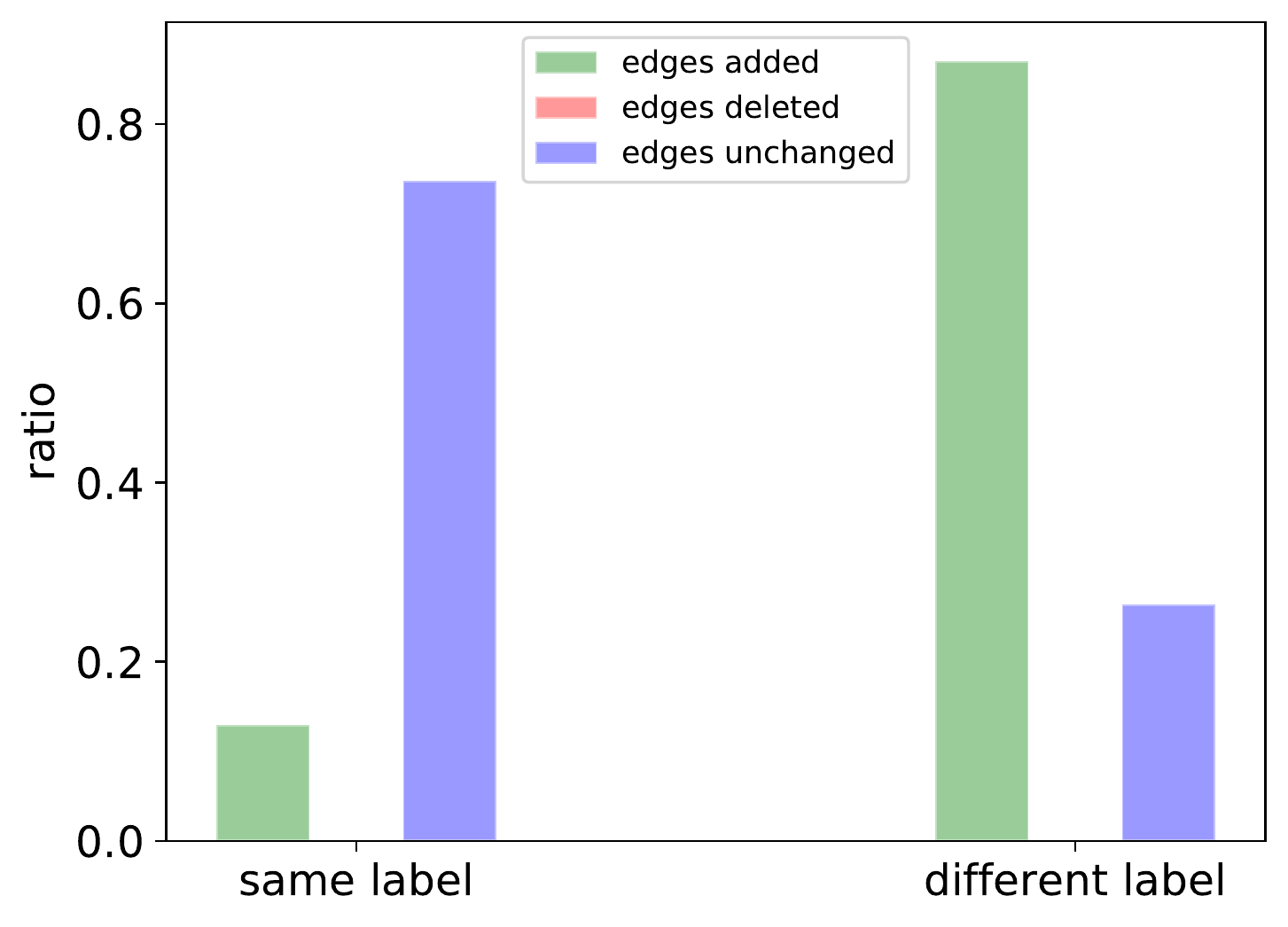}}
	  \subfigure[Polblogs] {\includegraphics[scale=0.33]{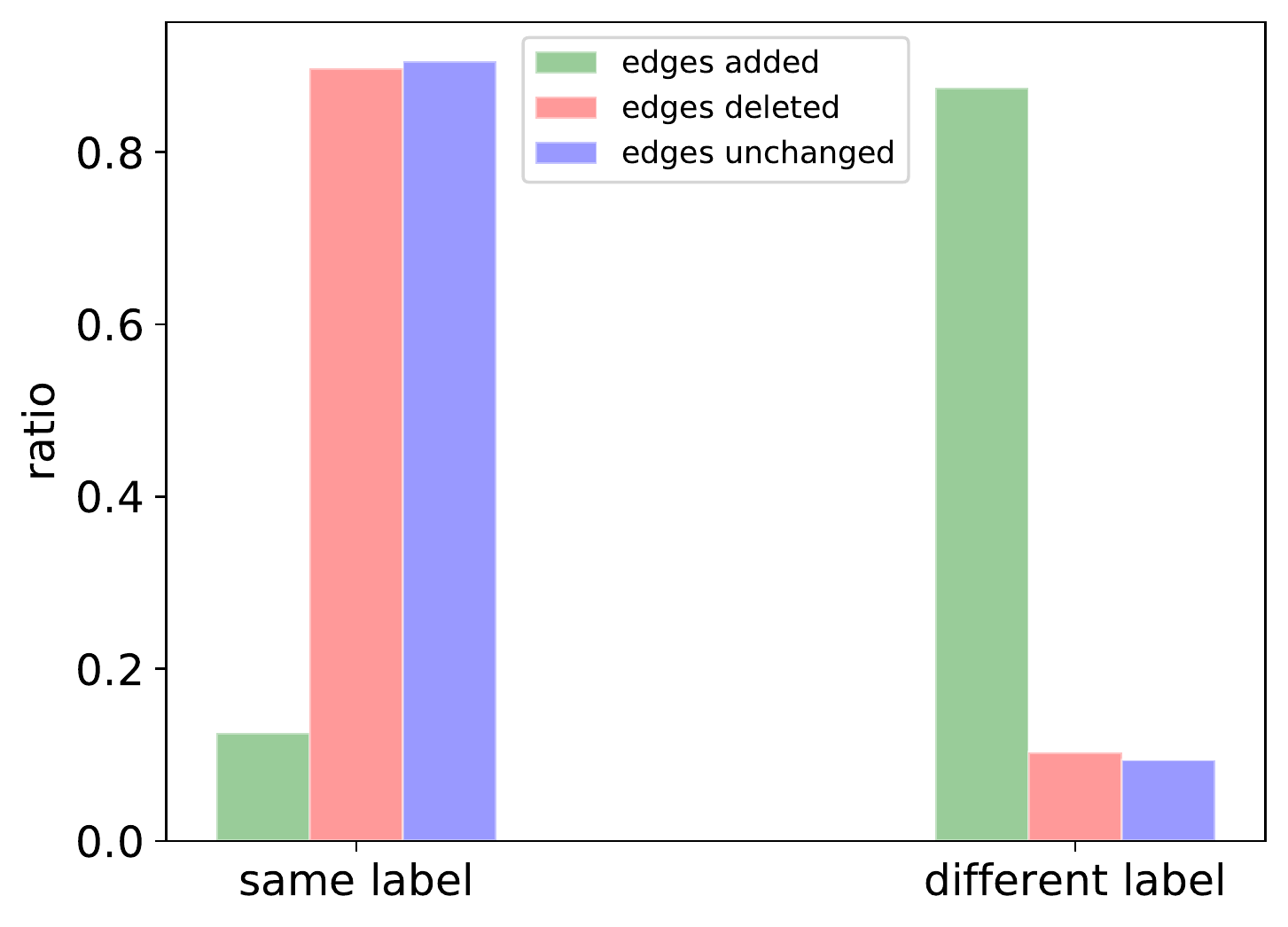}} 
	\end{center}
\caption{Label equality for Topology attack.}
\label{fig:labequality-top}
\end{figure*}

\begin{figure*}[htbp]
	\begin{center}
	  \subfigure[Cora] {\includegraphics[scale=0.33]{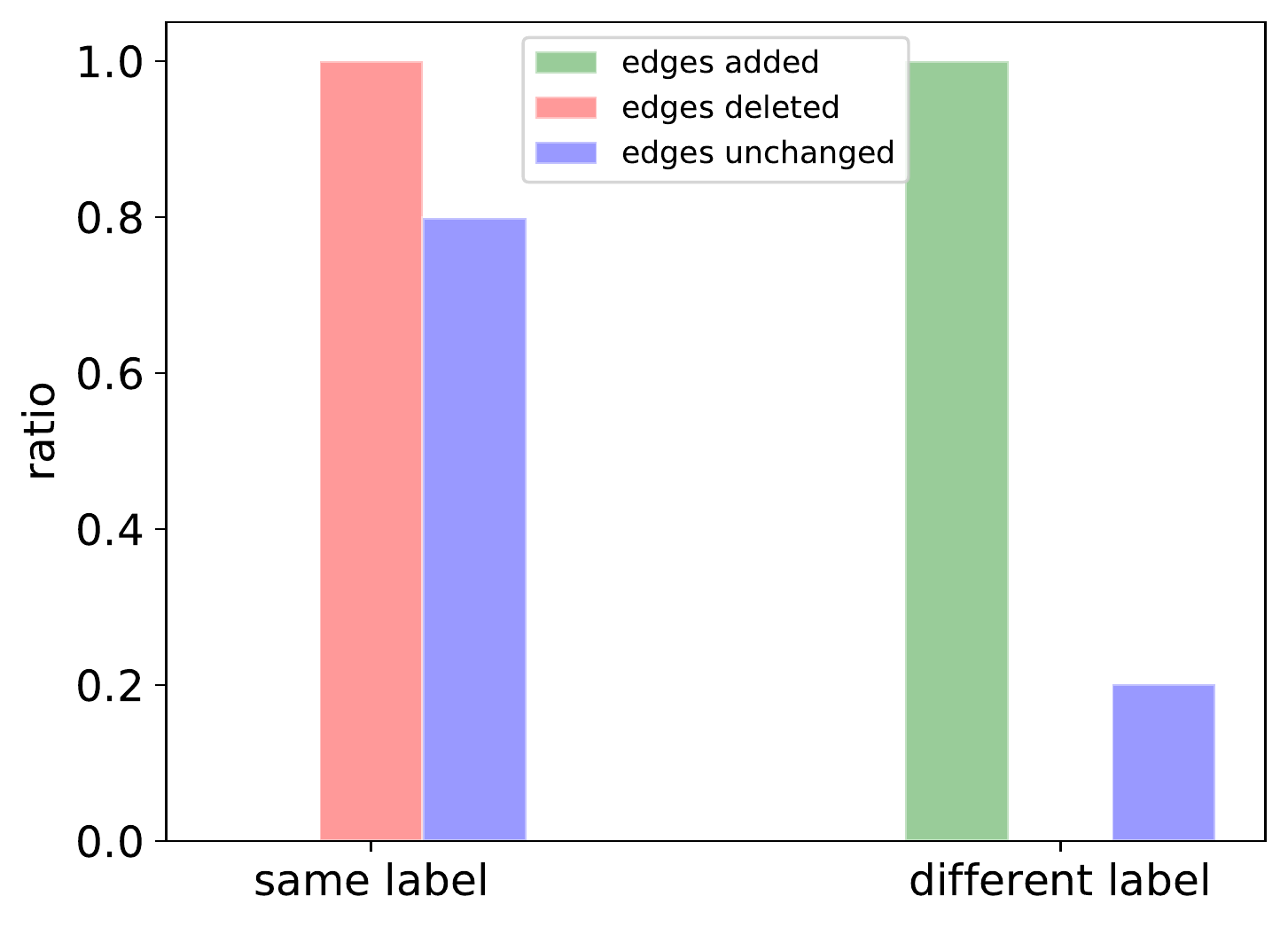}}
	  \subfigure[Citeseer]{\includegraphics[scale=0.33]{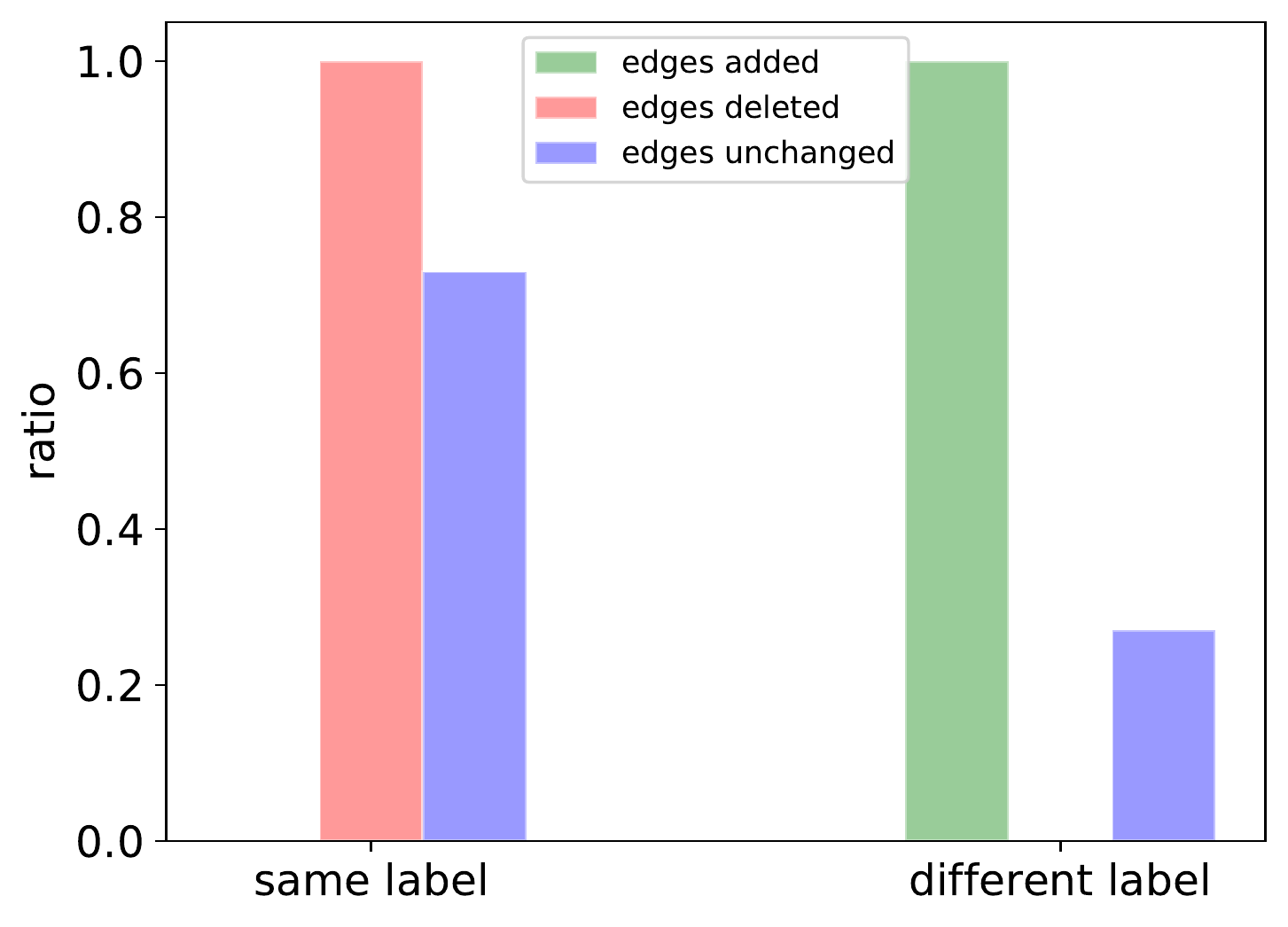}}
	  \subfigure[Polblogs] {\includegraphics[scale=0.33]{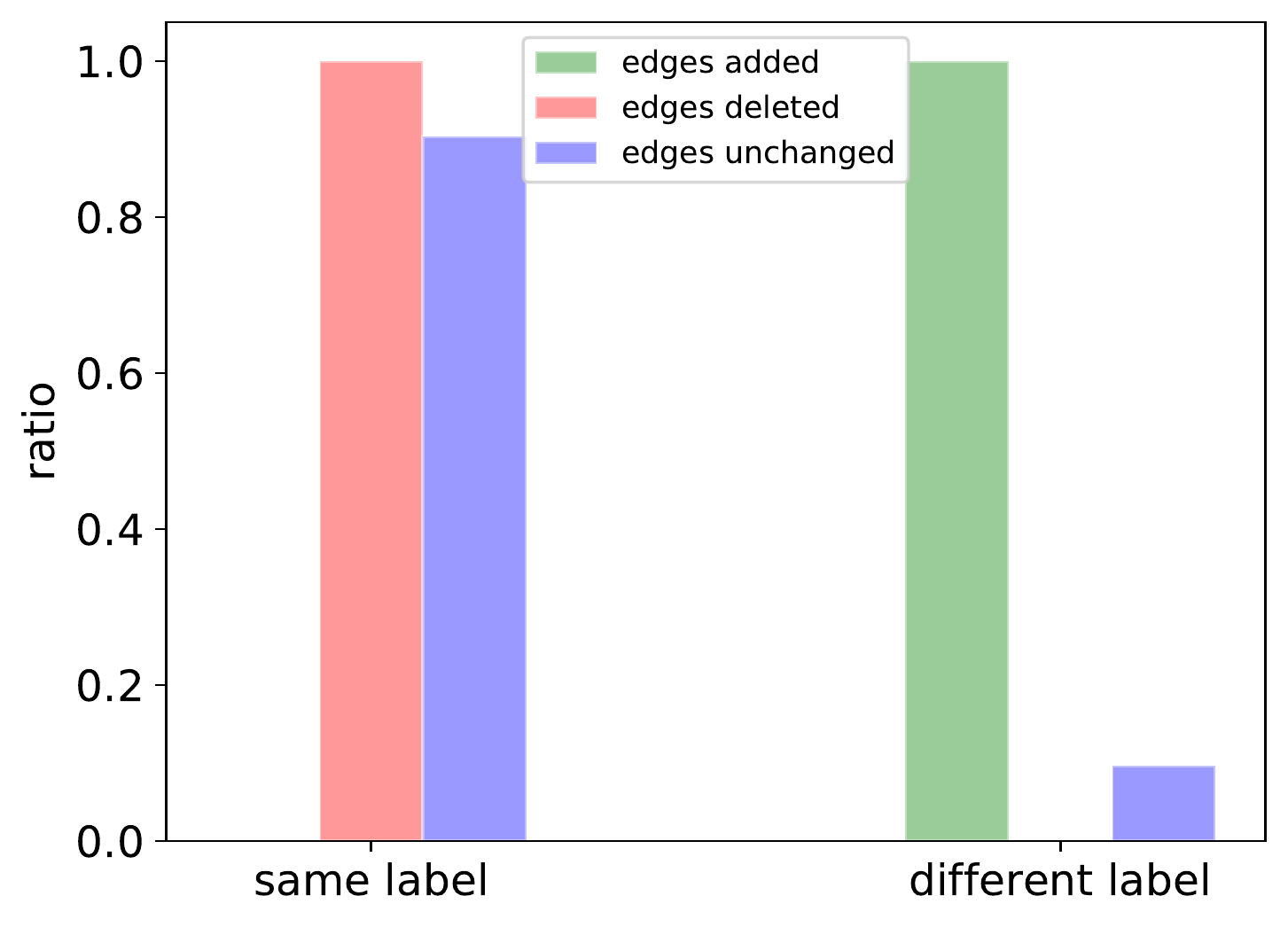}} 
	\end{center}
\caption{Label equality for DICE attack.}
\label{fig:labequality-dice}
\end{figure*}

The node feature similarity and label equality for Topology attack are shown in  Figure~\ref{fig:featuresimilarity-top} and Figure~\ref{fig:labequality-top}, respectively; the node feature similarity and label equality for DICE are shown in  Figure~\ref{fig:featuresimilarity-dice} and Figure~\ref{fig:labequality-dice}, respectively.

\vskip 6em

\vskip 3em

\subsection{Attack and Defense Performance for Topology Attack and DICE}
The attack and defense performance for DICE is shown in Table~\ref{tab:topology}; the performance for DICE is shown in Table~\ref{tab:dice}.
\begin{table}[h]
    \centering
    \caption{Performance (accuracy) under Topology attack.}
    \begin{threeparttable}
    \resizebox{0.47\textwidth}{!}{\begin{tabular}{@{}llllllll@{}}
    \toprule
    Dataset & $r$ (\%) & 0 & 5 & 10 & 15 & 20 & 25 \\
    \midrule
   
    \multirow{5}{*}{Cora}& GCN & 83.10 & 71.82 & 68.96 & 66.77 & 64.21 & 62.52 \\
    & Jaccard\footnote[1]{} & 82.39 &73.05 & 72.62 & 71.84 & 71.41 & 70.85\\
    & SVD\footnote[2]{} & 77.97 & 78.17 & 75.92 & 73.69 & 72.03 & 70.11 \\
    & RGCN & 84.81 & 72.68 & 71.15 & 69.38 & 67.92 & 67.23\\
    & GAT & 81.69 & 71.03 & 68.80 & 65.66 & 64.29 & 62.58\\
    \midrule
    
    \multirow{5}{*}{Citeseer} & GCN & 74.53 & 79.29  & 75.47 & 72.89 & 70.12 & 68.49 \\
    & Jaccard\footnote[1]{} & 74.82 &  79.07 & 76.76 & 74.29 & 71.87 & 69.55 \\
    & SVD\footnote[2]{} & 70.32 & 78.17 & 75.92 & 73.69 & 72.03 & 70.11\\
    & RGCN & 74.41 & 78.13 & 75.93 & 73.93 & 72.32 & 70.60\\
    & GAT & 74.23 & 77.52 & 74.09 & 71.90 & 69.62 & 66.99\\
    \midrule
  
    \multirow{5}{*}{Polblogs} & GCN & 95.80 & 72.04  & 65.87 & 63.35 & 61.06 & 58.49 \\
    
    & SVD\footnote[2]{} & 94.99 & 71.90 & 65.42 & 63.01 & 60.74 & 58.26\\
    & RGCN & 95.60 & 71.27 & 65.30 & 62.76 & 60.25 & 57.89\\
    & GAT & 95.40 & 72.56 & 65.97 & 63.35 & 60.94 & 58.77\\
    \bottomrule
    \end{tabular}}
    \begin{tablenotes}
    \item[1] Jaccard: GCN-Jaccard defense model. 
    \item[2] SVD: GCN-SVD defense model.
    \end{tablenotes}
\end{threeparttable}
\label{tab:topology}
\end{table}
\begin{table}[h]
    \centering
    \caption{Performance (accuracy) under DICE attack.}
    \begin{threeparttable}
    \resizebox{0.47\textwidth}{!}{\begin{tabular}{@{}cccccccc@{}}
    \toprule
    Dataset& $r$ (\%) & 0 & 5 & 10 & 15 & 20 & 25 \\
    \midrule
    \multirow{5}{*}{Cora}& GCN & 83.10 & 81.56 & 80.28 & 78.64 & 77.10 & 75.17\\
    & Jaccard\footnote[1]{} & 82.39 & 80.91 & 79.80 & 79.23 & 77.81 & 76.35\\
    & SVD\footnote[2]{} & 77.97 & 75.24 & 73.33 & 70.92 & 69.47 & 67.29 \\
    & RGCN & 84.81 & 83.53 & 82.56 & 80.70 & 79.30 & 77.89 \\
    & GAT & 81.69 & 78.86 & 77.50 & 74.56 & 70.54 & 69.25\\
    \midrule
    \multirow{5}{*}{Citeseer}& GCN & 74.53 & 74.41 & 73.61 & 71.98 & 71.47 & 69.92 \\
    & Jaccard\footnote[1]{} & 74.82 & 74.73 & 74.11 & 72.88 & 72.36 & 71.32 \\
    & SVD\footnote[2]{} & 70.32 & 70.68 & 69.89 & 68.59 & 67.09 &66.65 \\
    & RGCN & 74.41 &74.46 & 73.93 & 72.37 & 71.61 & 70.85 \\
    & GAT & 74.23 & 73.70 & 72.71 & 70.52 & 69.27 & 67.78\\
    \midrule
    \multirow{5}{*}{Polblogs}& GCN & 95.80 & 89.90 & 85.87 & 82.41 & 79.47 & 78.02 \\
    & SVD\footnote[2]{} & 94.99 & 90.88 & 87.79 &85.09 & 83.64 & 81.25\\
    & RGCN & 95.60 & 89.86 & 86.11 & 82.25 & 78.81 & 77.18 \\
    & GAT & 95.40 & 90.74 & 86.22 & 82.41 & 78.83 & 76.77 \\ 
    \bottomrule
    \end{tabular}}
    \begin{tablenotes}
    \item[1] Jaccard: GCN-Jaccard defense model. 
    \item[2] SVD: GCN-SVD defense model.
    \end{tablenotes}
\end{threeparttable}
\label{tab:dice}
\end{table}  

\vskip 4em

\section{Open Source Code}
We list some open source implementations of representative algorithms in Table~\ref{tab:code}.
\begin{table*}[!t]
\centering
\caption{A summary of open-source implementations}
\label{tab:code}
\begin{tabular}{@{}llll@{}}
\toprule
                         & Methods                                                                 & Framework                                                       & Github Link                                                                                                                                                        \\ \midrule
\multirow{12}{*}{Attack}  & PGD, Min-max~\cite{xu2019topology-attack}         & \begin{tabular}[c]{@{}l@{}}tensorflow\\ pytorch\end{tabular}    & \begin{tabular}[c]{@{}l@{}}https://github.com/KaidiXu/GCN\_ADV\_Train\\ https://github.com/DSE-MSU/DeepRobust\end{tabular}                                         \\ \cmidrule(l){2-4} 
                         & DICE~\cite{waniek2018hiding-dice}                 & python                                                          & https://github.com/DSE-MSU/DeepRobust                                                                                                                              \\ \cmidrule(l){2-4} 
                          & FGA~\cite{chen2018fast-gradient-network-embedding}                 & pytorch                                                          & https://github.com/DSE-MSU/DeepRobust                                                                                                                              \\ \cmidrule(l){2-4} 
                          & IG-FGSM~\cite{deep-insight-jaccard}                          & pytorch                                                         &     https://github.com/DSE-MSU/DeepRobust                                                                                                                                                               \\ \cmidrule(l){2-4} 
                          
                         & Nettack~\cite{nettack}                            & tensorflow                                                      & https://github.com/danielzuegner/nettack                                                                                                                           \\ \cmidrule(l){2-4} 
                         & Metattack~\cite{metattack}                        & \begin{tabular}[c]{@{}l@{}}tensorflow\\ pytorch\end{tabular}    & \begin{tabular}[c]{@{}l@{}}https://github.com/danielzuegner/gnn-meta-attack\\ https://github.com/ChandlerBang/pytorch-gnn-meta-attack\end{tabular}                 \\ \cmidrule(l){2-4} 
                         & RL-S2V~\cite{rl-s2v}                              & pytorch                                                         & https://github.com/Hanjun-Dai/graph\_adversarial\_attack                                                                                                           \\ \cmidrule(l){2-4} 
                         & Bojchevski et al.~\cite{node-embedding-poisoning}                        & tensorflow                                                      & https://github.com/abojchevski/node\_embedding\_attack                                                                                                             \\ \cmidrule(l){2-4} 
                         & GF-Attack~\cite{gf-attack}                        & tensoflow                                                       & https://github.com/SwiftieH/GFAttack                                                                                                                               \\ \midrule
\multirow{15}{*}{Defense} & RGCN~\cite{rgcn}                                  & \begin{tabular}[c]{@{}l@{}}tensorflow\\ \\ pytorch\end{tabular} & \begin{tabular}[c]{@{}l@{}}https://github.com/thumanlab/nrlweb/blob/master/static/\\ assets/download/RGCN.zip\\ https://github.com/DSE-MSU/DeepRobust\end{tabular} \\ \cmidrule(l){2-4} 
                         & GCN-Jaccard~\cite{deep-insight-jaccard}           & pytorch                                                         & https://github.com/DSE-MSU/DeepRobust                                                                                                                              \\ \cmidrule(l){2-4} 
                         & GCN-SVD~\cite{entezari2020all-svd}                & pytorch                                                         & https://github.com/DSE-MSU/DeepRobust                                                                                                                              \\ \cmidrule(l){2-4} 
                         & Pro-GNN~\cite{jin2020graph}                & pytorch                                                         & https://github.com/ChandlerBang/Pro-GNN                                                                                                                              \\ \cmidrule(l){2-4}
                         
& \begin{tabular}[c]{@{}l@{}}Adversarial \\  Training~\cite{xu2019topology-attack}\end{tabular}          & \begin{tabular}[c]{@{}l@{}}tensorflow\\ pytorch\end{tabular}    & \begin{tabular}[c]{@{}l@{}}https://github.com/KaidiXu/GCN\_ADV\_Train\\ https://github.com/DSE-MSU/DeepRobust\end{tabular}                                         \\ \cmidrule(l){2-4} 
                         & PA-GNN~\cite{pa-gnn}                              & tensorflow                                                      & https://github.com/tangxianfeng/PA-GNN                                                                                                                             \\ \cmidrule(l){2-4} 
                         & Graph-Cert~\cite{bojchevski2019certifiable-graph} & python                                                          & https://github.com/abojchevski/graph\_cert     \\ \cmidrule(l){2-4} 
                  & DefenseVGAE~\cite{zhang2020defensevgae} & pytorch                                                        & https://github.com/zhangao520/defense-vgae              \\ \cmidrule(l){2-4}     
                  & Bojchevski et al.~\cite{bojchevski2020efficient} & pytorch                                                          & https://github.com/abojchevski/sparse\_smoothing     \\ \cmidrule(l){2-4} 
                   & Z{\"u}gner et al.~\cite{zugner2020certifiable3} & python                                                        & https://github.com/danielzuegner/robust-gcn-structure              \\ \bottomrule
\end{tabular}
\end{table*}

\end{document}